\documentclass{article}

    \PassOptionsToPackage{numbers, compress}{natbib}
    \usepackage[preprint]{neurips_2021}

\usepackage[utf8]{inputenc} % allow utf-8 input
\usepackage[T1]{fontenc}    % use 8-bit T1 fonts
\usepackage{hyperref}       % hyperlinks
\usepackage{url}            % simple URL typesetting
\usepackage{booktabs}       % professional-quality tables
\usepackage{amsfonts}       % blackboard math symbols
\usepackage{nicefrac}       % compact symbols for 1/2, etc.
\usepackage{microtype}      % microtypography
\usepackage{xcolor}         % colors

\usepackage{amsmath,amsfonts,bm}
\usepackage{graphicx}
\usepackage[normalem]{ulem}
\useunder{\uline}{\ul}{}
\def\eqref#1{equation~\ref{#1}}
\def\1{\bm{1}}

\newcommand{\E}{\mathbb{E}}

\usepackage{amsthm,amsmath,amsfonts,bm,amssymb,bbm}
\usepackage{physics,graphicx}
\usepackage{xcolor}
\usepackage[shortlabels]{enumitem}

\newtheorem{prop}{Proposition}
\newtheorem{assumption}{Assumption}
\newtheorem{lemma}{Lemma}
\newtheorem{cor}{Corollary}

\DeclareMathOperator*{\argmax}{argmax}

\makeatletter
\newcommand*\rel@kern[1]{\kern#1\dimexpr\macc@kerna}
\newcommand*\widebar[1]{%
  \begingroup
  \def\mathaccent##1##2{%
    \rel@kern{0.8}%
    \overline{\rel@kern{-0.8}\macc@nucleus\rel@kern{0.2}}%
    \rel@kern{-0.2}%
  }%
  \macc@depth\@ne
  \let\math@bgroup\@empty \let\math@egroup\macc@set@skewchar
  \mathsurround\z@ \frozen@everymath{\mathgroup\macc@group\relax}%
  \macc@set@skewchar\relax
  \let\mathaccentV\macc@nested@a
  \macc@nested@a\relax111{#1}%
  \endgroup
}

\newsavebox{\mybox}\newsavebox{\mysim}
\newcommand{\distas}[1]{%
  \savebox{\mybox}{\hbox{\kern3pt$\scriptstyle#1$\kern3pt}}%
  \savebox{\mysim}{\hbox{$\sim$}}%
  \mathbin{\overset{#1}{\kern\z@\resizebox{\wd\mybox}{\ht\mysim}{$\sim$}}}%
}

\makeatother

\newcommand{\reals}{\mathbb{R}}
\newcommand{\one}{\mathbbm{1}}
\newcommand\ind[1]{\one\left[{#1}\right]}

\newcommand{\bH}{\bar{H}}
\newcommand{\hk}{\hat{k}}

\newcommand{\sbra}{\left(}
\newcommand{\sket}{\right)}

\newcommand{\vpara}[1]{\vspace{0.05in}\noindent\textbf{#1 }}

\usepackage{booktabs}

\title{Adversarial Attack on Graph Neural Networks\\ as An Influence Maximization Problem}

\author{%
    Jiaqi Ma \thanks{School of Information, University of Michigan, Ann Arbor, Michigan, USA} \thanks{Equal contribution.}\\
    \texttt{jiaqima@umich.edu} \\
    \And
    Junwei Deng \footnotemark[1] \footnotemark[2]\\
    \texttt{junweid@umich.edu} \\
    \And
    Qiaozhu Mei\footnotemark[1] \thanks{Department of EECS, University of Michigan, Ann Arbor, Michigan, USA} \\
    \texttt{qmei@umich.edu} \\
}

\begin{document}

\maketitle

\begin{abstract}
Graph neural networks (GNNs) have attracted increasing interests. With broad deployments of GNNs in real-world applications, there is an urgent need for understanding the robustness of GNNs under adversarial attacks, especially in realistic setups. In this work, we study the problem of attacking GNNs in a restricted and realistic setup, by perturbing the features of a small set of nodes,  with no access to model parameters and model predictions. Our formal analysis draws a connection between this type of attacks and an influence maximization problem on the graph. This connection not only enhances our understanding on the problem of adversarial attack on GNNs, but also allows us to propose a group of effective and practical attack strategies. Our experiments verify that the proposed attack strategies significantly degrade the performance of three popular GNN models and outperform baseline adversarial attack strategies. 
\end{abstract}

\section{Introduction}
There has been a surge of research interest recently in graph neural networks (GNNs)~\citep{wu2020comprehensive}, a family of deep learning models on graphs, as they have achieved superior performance on various tasks such as traffic forecasting~\citep{yu2017spatio}, social network analysis~\citep{li2017deepcas}, and recommender systems~\citep{ying2018graph,fan2019graph}. Given the successful applications of GNNs in online Web services, there are increasing concerns regarding the robustness of GNNs under adversarial attacks, especially in realistic scenarios. In addition, the research about adversarial attacks on GNNs in turns helps us better understand the intrinsic properties of existing GNN models. Indeed, there have been a line of research investigating various adversarial attack scenarios for GNNs~\citep{zugner2018adversarial,zugner_adversarial_2019,dai2018adversarial,bojchevski2018adversarial,ma2020black}, and many of them have been shown to be, unfortunately, vulnerable in these scenarios. In particular, \citet{ma2020black} examine an extremely restricted black-box attack scenario where the attacker has access to neither model parameters nor model predictions, yet they demonstrate that a greedy adversarial attack strategy can significantly degrade GNN performance due to the natural inductive biases of GNN binding to the graph structure. This scenario is motivated by real-world GNN applications on social networks, where attackers are only able to manipulate a limited number of user accounts, and they have no access to the GNN model parameters or predictions for the majority of users. 

In this work, we study adversarial attacks on GNNs under the aforementioned restricted and realistic setup. Specifically, an attack in this scenario is decomposed into two steps: 1) select a small set of nodes to be perturbed; 2) alter the node features according to domain knowledge up to a per-node budget. The focus of the study lies on the node selection step, so as in \citet{ma2020black}. The existing attack strategies, although empirically effective, are largely based on heuristics. We instead formulate the adversarial attack as an optimization problem to maximize the mis-classification rate over the selected set of nodes, and we carry out a formal analysis regarding this optimization problem. The proposed optimization problem is combinatorial and seems hard to solve in its original form. In addition, the mis-classification rate objective involves model parameters which are unknown in the restricted attack setup. We mitigate these difficulties by rewriting the problem and connecting it with \emph{influence maximization} on a variant of \emph{linear threshold model}\footnote{Strictly speaking, the diffusion model we connect with is slightly different from the standard linear threshold model, and should be called a \emph{general threshold model}~\citep{granovetter1978threshold,mossel2007submodularity}. But we view it as a (variant of) linear threshold model to avoid redundant notations.}~\citep{kempe2003maximizing} related to the original graph structure. 
Inspired by this connection, we show that, under certain distributional assumptions about the GNN, the expected mis-classification rate is submodular with respect to the selected set of nodes to perturb. The expected mis-classification rate is independent of the model parameters and can be efficiently optimized by a greedy algorithm thanks to its submodularity. Therefore, by specifying concrete distributions, we are able to derive a group of practical attack strategies maximizing the expected mis-classification rate. The connection with influence maximization also provides us nice interpretations regarding the problem of adversarial attack on GNNs. 

To empirically verify the effectiveness of the theory, we implement two adversarial attack strategies and test them on three popular GNN models, Graph Convoluntioal Network~(GCN)~\citep{kipf2016semi}, Graph Attention Network~(GAT)~\citep{velivckovic2018graph}, and Jumping Knowledge Network~(JKNet)~\citep{xu2018representation} with common benchmark datasets. Both attack strategies significantly outperform baseline attack strategies in terms of decreasing model accuracy. Finally, we summarize the contributions of our study as follows.
\begin{enumerate}[noitemsep,nolistsep]
    \item We formulate the problem of adversarial attack on GNNs as an optimization problem to maximize the mis-classification rate.
    \item We draw a novel connection between the problem of adversarial attacks on GNNs and that of influence maximization on a general threshold model. This connection helps us develop effective adversarial attack strategies under a restricted and realistic attack setup, and provides interpretations regarding the adversarial attack problem.
    \item We implement two variants of the proposed attack strategies and empirically demonstrate their effectiveness on popular GNNs.
\end{enumerate}

\section{Related Work}
There has been increasing research interest in adversarial attacks on GNNs recently. Detailed expositions of existing literature are made available in a couple of survey papers~\citep{jin2020adversarial,sun2018adversarial}. Given the heterogeneous nature of diverse graph structured data, there are numerous adversarial attack setups for GNN models. Following the taxonomy provided by \citet{jin2020adversarial}, the adversarial attack setup can be categorized based on (but not limited to) the machine learning task, the goal of the attack, the phase of the attack, the form of the attack, and the model knowledge that attacker has access to. First, there are two common types of tasks, node-level classification~\citep{zugner2018adversarial,dai2018adversarial,wu2019adversarial,entezari2020all} and graph-level classification~\citep{tang2020adversarial,dai2018adversarial}. The goal of the attack can be changing the predictions of a small and specific set of nodes (targeted attack)~\citep{zugner2018adversarial,dai2018adversarial} or degrading the overall GNN performance (untargeted attack)~\citep{zugner_adversarial_2019,sun2019node}. The attack can happen at the model training phases (poisoning attack)~\citep{zugner_adversarial_2019,sun2019node} or after training completes (evasion attack)~\citep{dai2018adversarial,chang2020restricted}. The form of the attack could be perturbing the node features~\citep{zugner2018adversarial,ma2020black} or altering the graph topology~\citep{dai2018adversarial,sun2019node}. Finally, depending on the knowledge (e.g. model parameters, model predictions, features, and labels, etc.) the attacker has access to, the attacks can be roughly categorized into white-box attacks~\citep{xu2019topology}, grey-box attacks~\citep{zugner2018adversarial,sun2019node}, or black-box attacks~\citep{dai2018adversarial,chang2020restricted,ma2020black}. However, it is worth noting that the borders of these three categories are blurry. In particular, the definition of a ``black-box'' attack varies much across literature.

Overall, the setup of interest in this paper can be categorized as \emph{node-level, untargeted, evasional, and black-box} attacks by perturbing the node features. While each setup configuration might find its suitable application scenarios, we believe that the black-box setups are particularly important as they are associated with many realistic scenarios. Among the existing studies on node-level black-box attacks, most of them~\citep{bojchevski2018adversarial,chang2020restricted,dai2018adversarial} still allow access to model predictions or some internal representations such as node embeddings. In this paper, we follow the most strict black-box setup~\citep{ma2020black} to our knowledge, which prohibits any probing of the model\footnote{A similar setting in the computer vision literature~\citep{chen2017zoo,li2020practical} is sometimes referred as a ``no-box'' attack to emphasize that neither model parameters nor model predictions can be accessed.}. Compared to \citet{ma2020black}, we develop attack strategies by directly analyzing the problem of maximizing mis-classification rate, rather than relying on heuristics.

\section{Preliminaries}
\subsection{Notations}
\label{sec:notation}
We start by introducing notations that will be used across this paper. Suppose we have an attributed graph $G=(V, E, X, y)$, where $V=\{1,2,\cdots,N\}$ is the set of $N$ nodes, $E\subseteq V\times V$ is the set of edges, $X\in \reals^{N\times D}$ is the node feature matrix with $D$-dimensional features, and $y\in \{1, 2, \cdots, K\}^N$ is the node label vector with $K$ classes. 

We denote a random walk transition matrix on the graph as $M\in \reals^{N\times N}$. For any $1\le i,j\le N$,
\begin{align*}
    M_{ij}=\left\{
\begin{array}{rcl}
\frac{1}{|\mathcal{N}_i|},      &      & {\rm if}\ (i, j)\in E \text{ or } i=j,\\
0,    &      & {\rm otherwise,}
\end{array} \right.
\end{align*}
where $\mathcal{N}_i=\{j\in V \mid (i,j)\in E \} \cup \{i\}$ is the set of neighbors of node $i$, including itself.

To ease the notation, for any matrix $A\in \reals^{D_1\times D_2}$ in this paper, we refer $A_j$ to the transpose of the $j$-th row of the matrix, i.e., $A_j \in \reals^{D_2}$. 

We consider a GNN model $f:\reals^{N\times D}\rightarrow \reals^{N\times K}$ that maps from the node feature matrix $X$ to the output logits of all nodes (denoted as $H\triangleq f(X) \in \reals^{N\times K}$). We assume the GNN $f$ has $L$ layers, with the $l$-th layer ($0 < l < L$) at node $i$ taking the form 
\[H_i^{(l)} = \text{ReLU}\sbra \sum_{j\in \mathcal{N}_i} \alpha_{ij} W^{(l)} H_j^{(l-1)} \sket,\]
where $W^{(l)}$ is the learnable weight matrix, $\text{ReLU}(\cdot)$ is an element-wise ReLU activation function, and different GNNs have different normalization terms $\alpha_{ij}$. We also define $H^{(0)} = X$ and $H = H^{(L)} = \sum_{j\in \mathcal{N}_i} \alpha_{ij} W^{(L)} H_j^{(L-1)}$.  Later in Section~\ref{sec:analysis}, we carry out our analysis on a GCN model with $\alpha_{ij} = 1/|\mathcal{N}_i|$~\citep{hamilton2017inductive}.

\subsection{The Adversarial Attack Setup}
\label{sec:setup}
Next we briefly introduce the adversarial attack setup proposed by \citet{ma2020black}. The goal of the attack is to perturb the node features of a few carefully selected nodes such that the model performance is maximally degraded. The attack is decomposed into two steps. 

In the first step (\emph{node selection step}), the attacker selects a set of nodes $S\subseteq V$ to be perturbed, under two constraints: 
\[|S| \le r \text{ and } |\mathcal{N}_i|\le m, \forall i\in S,\]
for some $0< r \ll N$ and $0 < m \ll \max_i |\mathcal{N}_i|$. These two constraints prevent the attacker from manipulating a lot of nodes or very important nodes as measured by the node degree, which makes the setup more realistic. 

In the second step (\emph{feature perturbation step}), the attacker is allowed to add a small constant perturbation $\epsilon\in \reals^D$ to the feature $X_i$ of each node $i\in S$ and obtain the perturbed feature $X'_i$, i.e.,
\[X'_i \triangleq X_i + \epsilon.\]
The perturbation vector $\epsilon$ is constructed based on the domain knowledge about the task but without access to the GNN model. For example, if the GNN model facilitates a recommender system for social media, an attacker may hack a handful of carefully selected users and manipulate their demographic features to get more other users exposed to certain political content the attacker desires. In practice, the perturbation vector $\epsilon$ can be tailored for different nodes being manipulated, given personalized knowledge about each node. But following \citet{ma2020black}, we consider the worst case where no personalization is available and hence $\epsilon$ remains constant for each node $i\in S$. 

\subsection{Influence Maximization on A Linear Threshold Model}
\label{sec:ltm}
Given an information/influence diffusion model on a social network, influence maximization is the problem of finding a small seed set of users such that they spread the maximum amount of influence over the network. In a linear threshold model~\citep{kempe2003maximizing}, the influence among nodes is characterized by a weighted directed adjacency matrix $I\in \reals^{N\times N}$ where $I_{ij}\ge 0$ for each $(i,j)\in E$ and $I_{ij}=0$ for each $(i,j)\notin E$. Given a seed set of nodes being activated at initial state, the influence passes through the graph to activate other nodes. There is a threshold vector $\eta\in \reals^N$ associated with the nodes, indicating the threshold of influence each node must have received from its active neighbors before it becomes activated. In particular, when the influence propagation comes to a stationary point, a node $i$ outside the seed set will be activated if and only if
\begin{align}
    \sum_{j\in \mathcal{N}_i, j\text{ is activated}} I_{ij} \ge \eta_i. \label{eq:threshold}
\end{align}

\section{Analysis of the Adversarial Attack Problem}
\label{sec:analysis}
In this section, we develop principled adversarial attack strategies under the setup stated in Section~\ref{sec:setup}. 

\subsection{Node Selection for Mis-classification Rate Maximization}
Suppose an attacker wants to attack a well-trained $L$-layer GCN model $f$. Following the two-step attack procedure, the attacker first selects a valid node set $S\in C_{r,m}\triangleq\{T\subseteq V \mid |T|\le r, |\mathcal{N}_i| \le m, \forall i\in T\}$ for some given constraints $r$ and $m$. Then the constant perturbation $\epsilon$ is added to the feature of each node in $S$, which leads to a perturbed feature matrix $X(S, \epsilon)$. Since our primary interest is the design of the node selection step, we shall omit $\epsilon$ and just write the perturbed feature as $X(S)$ for simplicity. We denote the output logits of the model after perturbation as $H(S) = f(X(S))$. Clearly, $H(\emptyset)$ equals to the matrix of output logits without attack.

In an untargeted attack, the attacker wants the model to make as many mistakes as possible, which is best measured by the mis-classification rate. Therefore we formulate the problem of selecting the node set as an optimization problem maximizing the mis-classification rate over $S$, with the two constraints quantified by $r, m$:
\begin{align}
    \max_{S\in C_{r,m}}\quad&\sum_{j=1}^N\ind{\max_{k=1,\cdots, K} H_{jk}(S)\neq H_{jy_j}(S)}, \label{eq:opt_original}
\end{align}
where $\ind{\cdot}$ is the indicator function. We drop normalizing constant $1/N$ in mis-classification rate.

At the first glance, the optimization problem~(\ref{eq:opt_original}) is a combinatorial optimization problem with a complicated objective function involving neural networks. In the following section, we demonstrate that, under a simplifying assumption, it can be connected to the influence maximization problem. 

\subsection{Connection to the Influence Maximization on A Linear Threshold Model}
\label{sec:connection}
\begin{figure*}
    \centering
    \includegraphics[width=0.7\textwidth]{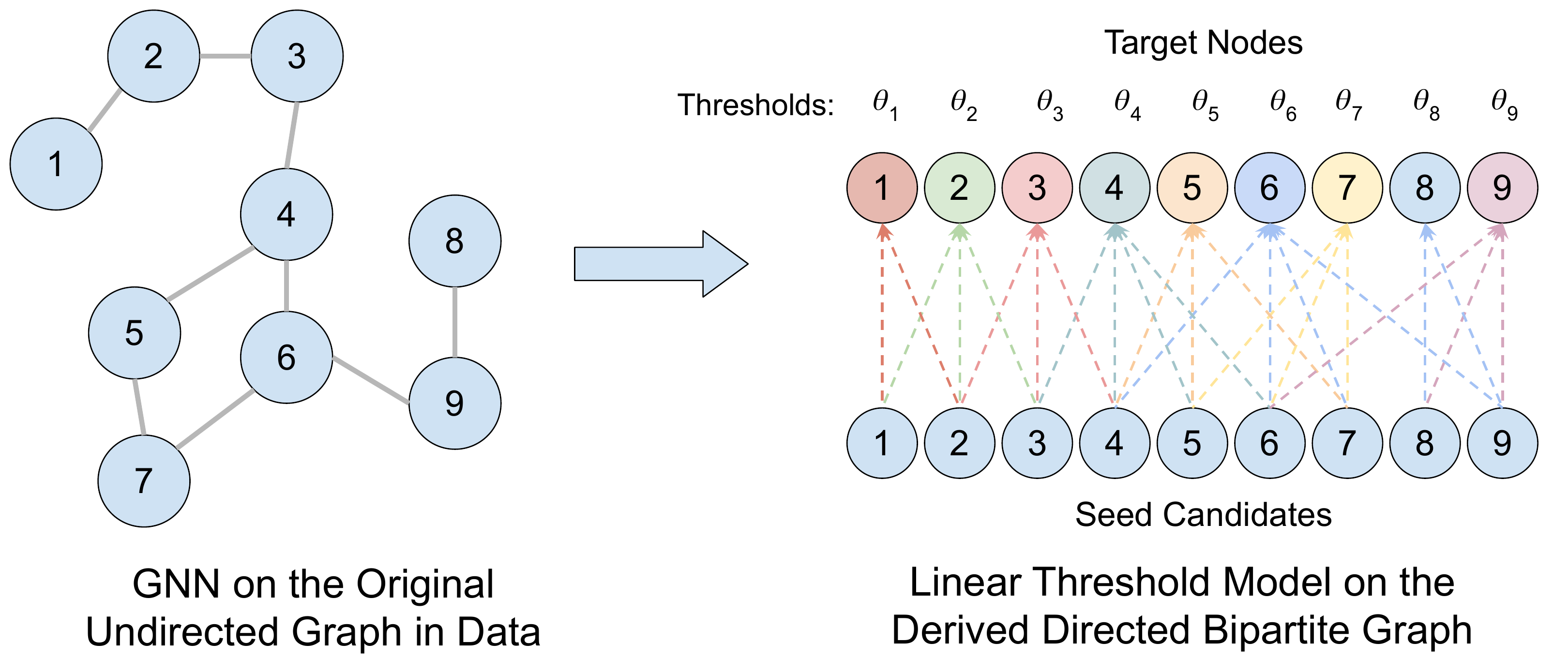}
    \caption{An illustrative example of the linear threshold model on the derived directed bipartite graph. To simplify the visualization, the GNN is assumed to have 1 layer, and therefore the derived directed bipartite graph have links from its zero-th (itself) and first order neighbors in the original graph. For a GNN with $k$ layers, the derived directed bipartite graph will have links from all its $l$-th order neighbors in the original graph, for any $0\le l\le k$. Each target node $i$ has its own threshold $\theta_i$ to be influenced (mis-classified). The edge weight depends on the random walk transition from the seed node to the target node.}
    \label{fig:ltm}
    \vskip -10pt
\end{figure*}

We first introduce a simplifying assumption of ReLU that has been widely used to ease the analysis of neural networks~\citep{choromanska2015loss,kawaguchi2016deep}, including GCN~\citep{xu2018representation}.

\begin{assumption} [\citet{xu2018representation}]
\label{assum:relu}
All the ReLU activations activate independently with the same probability, which implies that all paths in the computation graph of the GCN model are independently activated with the same probability of success $\rho$.
\end{assumption}

Under Assumption~\ref{assum:relu}, we are able to define $\bH(S) \triangleq \E_{\text{path}}\left[H(S)\right]$ for any $S\subseteq V$, where $\E_{\text{path}}\left[H(S)\right]$ indicates the expectation of $H(S)$ over the random activations of ReLU functions in the model. Then we can rewrite problem~(\ref{eq:opt_original}) in a form that is similar to the influence maximization objective on a linear threshold model. The influence weight matrix is defined by the $L$-step random walk transition matrix $B\triangleq M^L$. And the threshold for each node is related to the original output logits $\bH(\emptyset)$, the perturbation vector $\epsilon$, and the product of the GCN weights $W\triangleq \rho\cdot\prod_{l=L}^1 W^{(l)} \in \reals^{K\times D}$. Formally, we have the following Proposition~\ref{prop:opt_ltm}, whose proof can be found in Appendix~\ref{appendix:proof-opt_ltm}.
\begin{prop}
\label{prop:opt_ltm}
If we replace $H(\cdot)$ by $\bH(\cdot)$ in problem~(\ref{eq:opt_original}), then we can rewrite the optimization problem as follows,
\begin{align}
    \max_{S\in C_{r,m}}\quad&\sum_{j=1}^N\ind{\sum_{i\in S}B_{ji} > \theta_j}, \label{eq:opt_ltm}
\end{align}
where, for $\hk_j=\argmax_{k=1,\cdots,K}\bH_{jk}(S)$,
\begin{align}
    \theta_j \triangleq \frac{\bH_{jy_j}(\emptyset) - \bH_{j\hk_j}(\emptyset)}{(W_{\hk_j} - W_{y_j})^T\epsilon}. \label{eq:theta}
\end{align}
In particular, if $\hat{k}_j = y_j$, we define $\theta_j=\infty$.
\end{prop}

\vpara{Interpretations of the new objective~(\ref{eq:opt_ltm}).} The new optimization objective~(\ref{eq:opt_ltm}) has nice interpretations. The $L$-step random walk transition matrix measures the pairwise influence from input nodes to target nodes in the GCN model and $\sum_{i\in S}B_{ji}$ can be viewed as measuring the influence of nodes in $S$ on a target nodes $j$. In each $\theta_j$, the numerator $\bH_{jy_j}(\emptyset) - \bH_{j\hk_j}(\emptyset)$ can be viewed as the logit margin between the correct class and those wrong classes, which measures the robustness of the prediction on node $j$. The denominator $(W_{\hk_j} - W_{y_j})^T\epsilon$ measures how effective the perturbation is. In combination, $\theta_j$ measures how difficult it is to mis-classify the node $j$ with perturbation $\epsilon$. This new objective nicely separates the influence between nodes and the node-specific robustness.

Note the form of each term inside the summation over $N$ in Eq.~(\ref{eq:opt_ltm}), $\ind{\sum_{i\in S}B_{ji} > \theta_j}$, is very similar to that of Eq.~(\ref{eq:threshold}). In fact, the objective~(\ref{eq:opt_ltm}) can be viewed as the influence maximization objective on a directed bipartite graph derived from the original graph, as shown in Figure~\ref{fig:ltm}. The derived bipartite graph has $N$ nodes on both sides (assuming we call them the seed candidate side $\mathcal{S}$ and target node side $\mathcal{T}$), and there are edges pointing from side $\mathcal{S}$ to side $\mathcal{T}$ but not the converse way. The edge weight from the node $i$ on the side $\mathcal{S}$ to the node $j$ on the side $\mathcal{T}$ ($1\le i,j\le N$) is defined as $B_{ji}$. Then it is easy to see that the problem~(\ref{eq:opt_ltm}) is equivalent to the influence maximization problem on the bipartite graph with the node-specific thresholds being $\theta_j, j=1,\cdots,N$. 

\vpara{Two difficulties for solving the problem~(\ref{eq:opt_ltm}).} While we now have got better interpretations of the original mis-classification rate maximization problem in terms of influence maximization, we still face two major difficulties before we can develop an algorithm to solve the problem. The first difficulty is that we do not known the value of $\theta$ in a black-box attack setup as it involves the model parameters. The second difficulty is that, even if $\theta$ is given, influence maximization on the seemingly simple bipartite graph is still NP-hard, as we show in Lemma~\ref{lemma:np} (see Appendix~\ref{appendix:proof-np} for the proof).
\begin{lemma}
\label{lemma:np}
The influence maximization problem on a directed bipartite graph with linear threshold model is NP-hard.
\end{lemma}

\subsection{Assumptions on the Thresholds}
Next, we mitigate the aforementioned two difficulties by making distribution assumptions on the thresholds $\theta$.

It is well-known that if the threshold $\theta_j$ of each node $j$ is drawn uniformly at random from the interval $[0, 1]$, the expected objective of a general linear threshold model is submodular, which leads to an efficient greedy algorithm that solves the expected influence maximization problem with a performance guarantee~\citep{kempe2003maximizing}. In light of this fact regarding the general linear threshold model, we show (in Proposition~\ref{prop:submodular}) that a mild assumption on the distribution of $\theta$ will guarantee the expectation of the objective~(\ref{eq:opt_ltm}) to be submodular, thanks to the simple bipartite structure. The proof of Proposition~\ref{prop:submodular} can be found in Appendix~\ref{appendix:proof-submodular}.
\begin{prop}
\label{prop:submodular}
Suppose the individual thresholds are random variables drawn from some distributions, and the marginal cumulative distribution function of the threshold $\theta_j$ for node $j$ is $F_j$, $j=1,\cdots,N$. If $F_1, \cdots, F_N$ are individually concave in the domain $[0, +\infty)$, then the expectation of the objective~(\ref{eq:opt_ltm}),
\begin{align}
    h(S)\triangleq \E_{\theta_1,\cdots, \theta_N}\sum_{j=1}^N\ind{\sum_{i\in S}B_{ji} > \theta_j}, \label{eq:opt_expected}
\end{align}
is submodular.
\end{prop}
Note that here we do not need the thresholds $\theta$ to be independent from each other, and we only require the marginal probability density function of each $\theta_j$ to be non-increasing on the positive region.

Proposition~\ref{prop:submodular} partially addresses the second difficulty. While we still do not have a solution to the original problem~(\ref{eq:opt_ltm}), we now know that for a wide range of distributions of $\theta$, the expected mis-classification rate is submodular and can be approximated efficiently through a greedy algorithm. 

For the first difficulty, we propose to explicitly specify a simple distribution for $\theta$ and optimize the expected mis-classification rate $h(S)$, which no longer involves any model parameters and gives us a black-box attack strategy. While this seems to radically deviate from the original optimization objective~(\ref{eq:opt_ltm}), in the following Section~\ref{sec:exp}, we empirically show that we only need a crude characterization of the distribution of $\theta$ to obtain effective attack strategies.

\vpara{The concrete black-box attack strategies.} Below we derive two concrete black-box attack strategies by specifying the distribution of $\theta$ to be uniform distributions and normal distributions respectively. 

\begin{cor}
\label{cor:unif}
If $a, b > 0$ and $\theta_j\distas{i.i.d.}\text{ uniform }(-b, a)$, then 
\begin{align}
    h(S) = \frac{1}{a+b} \sum_{j=1}^N \left(\min(\sum_{i\in S}B_{ji}, a) + b\right), \label{eq:opt_unif}
\end{align}
and $h(S)$ is submodular.
\end{cor}

\begin{cor}
\label{cor:normal}
If $\sigma > 0$ and $\theta_j\distas{i.i.d.}\mathcal{N}(0, \sigma^2)$, then 
\begin{align}
    h(S) = \frac{1}{2} \sum_{j=1}^N \left(1+ \text{erf}\left(\frac{\sum_{i\in S}B_{ji}}{\sigma\sqrt{2}}\right)\right), \label{eq:opt_normal}
\end{align}
where $\text{erf}(\cdot)$ is the Gauss error function. And $h(S)$ is submodular.
\end{cor}

Corollary~\ref{cor:unif} and~\ref{cor:normal} follow directly from Proposition~\ref{prop:submodular} given the cumulative distribution functions of the uniform distribution and the normal distribution as well as the fact that they are concave at the positive region. In particular, Eq.~(\ref{eq:opt_unif}) belongs to a well-known submodular function family named the \textit{saturated coverage function}~\citep{lin2011class,iyer2015submodular}. Under assumptions in Corollary~\ref{cor:unif}, the adversarial attack problem reduces to the classic influence maximization problem under the linear threshold model where the thresholds follow uniform distributions.

We name the attack strategies obtained by greedily maximizing the objectives (\ref{eq:opt_unif}) and (\ref{eq:opt_normal}) as \textbf{InfMax-Unif} and \textbf{InfMax-Norm} respectively. Specifically, each strategy iteratively selects nodes into the set to be perturbed up to a given size. At each iteration, the node, combining with the existing set, that maximizes Eq.~(\ref{eq:opt_unif}) or Eq.~(\ref{eq:opt_normal}) will be selected. 

\subsection{Discussions on the Approximations}
\label{sec:disc}
From problem~(\ref{eq:opt_ltm}) to our final attack strategies, we have made two major approximations to address the two difficulties that we raised at the end of Section~\ref{sec:connection}. 

The first approximation is we go from the original optimization problem to its expected version. Note that $\theta$ depends on both the model parameters and the data, which we do not have full access to. The first approximation treats them as random, and takes expectation over $\theta$, which integrates out the randomness in data and the model training process. And the resulted expected objective function $h(S)$ is submodular under the conditions in Proposition~\ref{prop:submodular}. A natural question regarding this approximation is how does the mis-classification rate (\ref{eq:opt_ltm}) concentrate around its expectation (\ref{eq:opt_expected})? If $\theta$ are independent, the indicator variables in (\ref{eq:opt_expected}) are also independent, and it is easy to show the mis-classification rate is well-concentrated for a large graph size $N$ through Hoeffding's inequality. However, the independence assumption is unrealistic in the case of GNN as the predictions of adjacent nodes should be correlated. Further note that $\theta$ can be written in terms of linear combinations of node features. With extra assumptions on the node features and the graph structure, one may be able to carry out finer analysis on the covariance of $\theta$, and thus how well the mis-classification rate concentrates. We leave this analysis for future work.

The second approximation is that we further specify simple distributions of $\theta$, which highly likely deviate much from the real distribution. On one hand, our superior empirical results shown in Section~\ref{sec:exp} suggest that these simple strategies are practical enough for some applications. On the other hand, this leaves room for further improvement in real-world scenarios if we have more knowledge regarding the distribution of $\theta$. For example, if an attacker has a very limited number of API calls to access the model predictions, these calls are 
likely insufficient to train a reinforcement-learning-based attack strategies but they can be effectively used to better estimate the distribution of $\theta$. 

\section{Experiments}
\label{sec:exp}

\begin{table*}
\caption{Summary of the attack performance in terms of test accuracy (\%), the lower the better attack. \textbf{Bold} denotes the best performing strategy in each setup. \uline{Underline} indicates our strategy outperforms all the baseline strategies. Asterisk (*) means the difference between our strategy and the best baseline strategy is statistically significant by a pairwise t-test at significance level 0.05. The error bar ($\pm$) denotes the standard error of the mean by 40 independent trials. We test on two setups of the node degree threshold, $m$, by setting it equal to the lowest degree of the top 10\% and 30\% nodes respectively.}
\label{tab:results}
\resizebox{\textwidth}{!}{%
\begin{tabular}{llllllllll}
\hline
\multicolumn{1}{l|}{}            & \multicolumn{3}{c|}{Cora}                                                                                       & \multicolumn{3}{c|}{Citeseer}                                                                                   & \multicolumn{3}{c}{Pubmed}                                                                 \\
\multicolumn{1}{l|}{Method}      & JKNetMaxpool                 & GCN                          & \multicolumn{1}{l|}{GAT}                          & JKNetMaxpool                 & GCN                          & \multicolumn{1}{l|}{GAT}                          & JKNetMaxpool                 & GCN                          & GAT                          \\ \hline
\multicolumn{1}{l|}{None}        & $85.9\pm0.1$                 & $85.5\pm0.2$                 & \multicolumn{1}{l|}{$87.7\pm0.2$}                 & $73.0\pm0.2$                 & $75.0\pm0.2$                 & \multicolumn{1}{l|}{$74.8\pm0.2$}                 & $85.7\pm0.1$                 & $85.7\pm0.1$                   & $85.2\pm0.1$                 \\ \hline
\multicolumn{10}{c}{Threshold 10\%}                                                                                                                                                                                                                                                                                                                                \\ \hline
\multicolumn{1}{l|}{Random}      & $70.5\pm1.2$                 & $77.6\pm0.4$                 & \multicolumn{1}{l|}{$71.5\pm1.0$}                 & $59.0\pm0.9$                 & $69.7\pm0.3$                 & \multicolumn{1}{l|}{$70.8\pm0.4$}                 & $74.7\pm0.9$                 & $79.1\pm0.3$                 & $72.8\pm1.0$                 \\
\multicolumn{1}{l|}{Degree}      & $63.9\pm1.3$                 & $73.4\pm0.4$                 & \multicolumn{1}{l|}{$65.9\pm1.2$}                 & $51.1\pm0.9$                 & $63.9\pm0.3$                 & \multicolumn{1}{l|}{$65.6\pm0.7$}                 & $61.2\pm1.5$                 & $73.2\pm0.6$                 & $62.2\pm1.5$                 \\
\multicolumn{1}{l|}{Pagerank}    & $71.6\pm0.9$                 & $75.7\pm0.3$                 & \multicolumn{1}{l|}{$72.7\pm0.8$}                 & $61.3\pm0.8$                 & $69.5\pm0.3$                 & \multicolumn{1}{l|}{$70.4\pm0.4$}                 & $73.7\pm0.8$                 & $78.7\pm0.2$                 & $73.3\pm0.6$                 \\
\multicolumn{1}{l|}{Betweenness} & $65.2\pm1.6$                 & $72.4\pm0.5$                 & \multicolumn{1}{l|}{$64.3\pm1.2$}                 & $50.8\pm1.0$                 & $65.6\pm0.3$                 & \multicolumn{1}{l|}{$67.1\pm0.6$}                 & $66.9\pm1.4$                 & $74.6\pm0.6$                 & $62.9\pm1.6$                 \\
\multicolumn{1}{l|}{RWCS}        & $75.0\pm0.6$                 & $75.7\pm0.3$                 & \multicolumn{1}{l|}{$69.6\pm0.9$}                 & $61.4\pm0.8$                 & $69.5\pm0.3$                 & \multicolumn{1}{l|}{$70.3\pm0.4$}                 & $73.0\pm0.8$                 & $78.1\pm0.4$                 & $71.5\pm0.8$                 \\
\multicolumn{1}{l|}{GC-RWCS}     & $57.6\pm2.0$                 & $70.0\pm0.6$                 & \multicolumn{1}{l|}{$57.5\pm1.5$}                 & $45.0\pm0.8$                 & \boldmath{$60.5\pm0.4$}        & \multicolumn{1}{l|}{$62.4\pm1.0$}                 & $56.7\pm2.0$                 & $70.4\pm0.8$                 & $55.4\pm2.2$                 \\
\multicolumn{1}{l|}{InfMax-Unif} & {\ul \boldmath{$56.5\pm2.1*$}} & {\ul \boldmath{$68.8\pm0.7*$}} & \multicolumn{1}{l|}{{\ul \boldmath{$55.9\pm1.6*$}}} & {\ul \boldmath{$44.4\pm0.8*$}} & $60.7\pm0.5$                 & \multicolumn{1}{l|}{$62.4\pm1.0$}                 & {\ul $55.6\pm2.0*$}          & {\ul $69.5\pm0.9*$}          & {\ul $53.7\pm2.3*$}          \\
\multicolumn{1}{l|}{InfMax-Norm} & {\ul $56.9\pm2.1*$}          & {\ul $69.1\pm0.7*$}          & \multicolumn{1}{l|}{{\ul $57.1\pm1.5*$}}          & $45.0\pm0.8$                 & \boldmath{$60.5\pm0.5$}        & \multicolumn{1}{l|}{{\ul \boldmath{$62.3\pm1.0$}}}  & {\ul \boldmath{$54.0\pm2.0*$}} & {\ul \boldmath{$67.5\pm1.0*$}} & {\ul \boldmath{$52.5\pm2.4*$}} \\ \hline
\multicolumn{10}{c}{Threshold 30\%}                                                                                                                                                                                                                                                                                                                                \\ \hline
\multicolumn{1}{l|}{Random}      & $73.0\pm0.9$                 & $78.9\pm0.3$                 & \multicolumn{1}{l|}{$74.1\pm0.9$}                 & $61.2\pm1.1$                 & $71.1\pm0.4$                 & \multicolumn{1}{l|}{$71.9\pm0.4$}                 & $77.0\pm0.6$                 & $80.0\pm0.3$                 & $75.8\pm0.8$                 \\
\multicolumn{1}{l|}{Degree}      & $68.8\pm1.2$                 & $76.3\pm0.4$                 & \multicolumn{1}{l|}{$69.2\pm1.0$}                 & $56.0\pm1.0$                 & $68.4\pm0.3$                 & \multicolumn{1}{l|}{$69.1\pm0.5$}                 & $70.9\pm1.1$                 & $76.9\pm0.5$                 & $69.8\pm1.2$                 \\
\multicolumn{1}{l|}{Pagerank}    & $77.2\pm0.5$                 & $80.5\pm0.3$                 & \multicolumn{1}{l|}{$81.9\pm0.4$}                 & $68.2\pm0.8$                 & $72.4\pm0.3$                 & \multicolumn{1}{l|}{$73.0\pm0.3$}                 & $80.3\pm0.3$                 & $81.5\pm0.2$                 & $79.6\pm0.3$                 \\
\multicolumn{1}{l|}{Betweenness} & $68.9\pm1.4$                 & $74.3\pm0.5$                 & \multicolumn{1}{l|}{$66.5\pm1.0$}                 & $54.1\pm1.1$                 & $67.9\pm0.3$                 & \multicolumn{1}{l|}{$68.7\pm0.6$}                 & $75.5\pm0.7$                 & $78.4\pm0.4$                 & $72.5\pm1.0$                 \\
\multicolumn{1}{l|}{RWCS}        & $77.1\pm0.5$                 & $80.6\pm0.3$                 & \multicolumn{1}{l|}{$81.9\pm0.4$}                 & $68.2\pm0.8$                 & $72.4\pm0.3$                 & \multicolumn{1}{l|}{$73.1\pm0.3$}                 & $78.5\pm0.3$                 & $80.0\pm0.2$                 & $77.8\pm0.4$                 \\
\multicolumn{1}{l|}{GC-RWCS}     & $63.5\pm1.8$                 & $72.1\pm0.5$                 & \multicolumn{1}{l|}{$61.5\pm1.3$}                 & $48.2\pm1.0$                 & $64.0\pm0.4$                 & \multicolumn{1}{l|}{$65.1\pm0.8$}                 & $70.8\pm1.0$                 & $75.9\pm0.6$                 & $69.1\pm1.3$                 \\
\multicolumn{1}{l|}{InfMax-Unif} & {\ul \boldmath{$62.0\pm2.0*$}} & {\ul \boldmath{$71.7\pm0.6$}}  & \multicolumn{1}{l|}{{\ul \boldmath{$59.1\pm1.4*$}}} & {\ul \boldmath{$47.2\pm0.9*$}} & {\ul \boldmath{$62.8\pm0.4*$}} & \multicolumn{1}{l|}{{\ul \boldmath{$64.6\pm0.9*$}}} & {\ul \boldmath{$66.3\pm1.6*$}} & {\ul $75.2\pm0.6*$}          & {\ul $64.1\pm1.8*$}          \\
\multicolumn{1}{l|}{InfMax-Norm} & {\ul $62.3\pm2.0*$}          & $72.1\pm0.6$                 & \multicolumn{1}{l|}{{\ul $59.5\pm1.4*$}}          & {\ul $47.3\pm0.9*$}          & {\ul \boldmath{$62.8\pm0.4*$}} & \multicolumn{1}{l|}{{\ul \boldmath{$64.6\pm0.8*$}}}         & {\ul $66.8\pm1.5*$}          & {\ul \boldmath{$74.5\pm0.7*$}} & {\ul \boldmath{$63.9\pm1.8*$}} \\ \hline
\end{tabular}%
}
\end{table*}

In this section, we empirically evaluate the performance of the proposed attack strategies (InfMax-Unif and InfMax-Norm) against several baseline attack strategies, closely following the experiment setup of \citet{ma2020black}. We also visualize the distributions of $\theta$ and have a case study of the selected nodes to gain better qualitative understandings of the proposed methods. Some additional experiments such as sensitivity analyses are provided in Appenix~\ref{appendix:additional}.

\subsection{Attack Strategies for Comparison}
\label{sec:baseline}
\vpara{Implementation of InfMax-Unif and InfMax-Norm.} For the proposed InfMax-Unif and InfMax-Norm, there are two hyper-parameters respectively to be specified. Recall $B=M^L$, the first hyper-parameter for both method is $L$. We set $L=4$ following RWCS and GC-RWCS. We note that, for the attack strategies to be effective in practice, the hyper-parameter $L$ does not have to be the same as the number of layers of the GNN being attacked, as we will show in the experiments. For InfMax-Unif, there are two additional distribution hyper-parameters $a, b$. However, $b$ does not influence the selection of nodes so we only need to specify $a$. For InfMax-Norm, we need to specify the distribution parameter $\sigma$. We fix $a=0.01$ and $\sigma=0.01$ across all the experiment setups. Theoretically, the optimal choice of $a$ or $\sigma$ should depend on the perturbation vector $\epsilon$ as well as the dataset. However, we find the proposed InfMax-Unif and InfMax-Norm strategies are fairly robust with respect to the choice of $a$ or $\sigma$ (see the sensitivity analysis in Appendix~\ref{appendix:sensitivity}).

\vpara{Baseline strategies.} We compare with five baseline strategies, \textbf{Degree}, \textbf{Betweenness}, \textbf{PageRank}, Random Walk Column Sum (\textbf{RWCS}), and Greedily-Corrected RWCS (\textbf{GC-RWCS}). We briefly introduce the baselines below and leave more details of them in Appendix~\ref{appendix:baseline}. 

The first three strategies, as suggested by their names, correspond to three node centrality scores. These strategies select nodes with the highest node centrality scores subject to the constraint $C_{r,m}$. 

RWCS and GC-RWCS are two black-box attack strategies proposed by \citet{ma2020black}. RWCS is derived by maximizing the cross-entropy classification loss but with certain approximations. In practice, RWCS has a simple form: selects nodes with highest importance scores defined as $I(i) = \sum_{j=1}^N[M^L]_{ji}$ (recall that $M$ is the random walk transition matrix). We set the hyper-parameter $L=4$ following \citet{ma2020black}. GC-RWCS further applies a few heuristics on top of RWCS to achieve better mis-classification rate. Specifically, it dynamically updates the RWCS importance score based on a heuristic. It also removes a local neighborhood of the selected node after selecting each node.  In the experiment, we set the hyper-parameters of GC-RWCS $L=4$, $l=30$, and $k=1$ as suggested in their original paper. Interestingly, RWCS can be viewed as a special case of InfMax-Unif if we set $a=\infty$ (or large enough). And GC-RWCS without removing the local neighborhood step can also be viewed a modified version of InfMax-Unif.

\begin{figure*}
    \centering
    \vskip -10pt
    \includegraphics[width=0.9\textwidth]{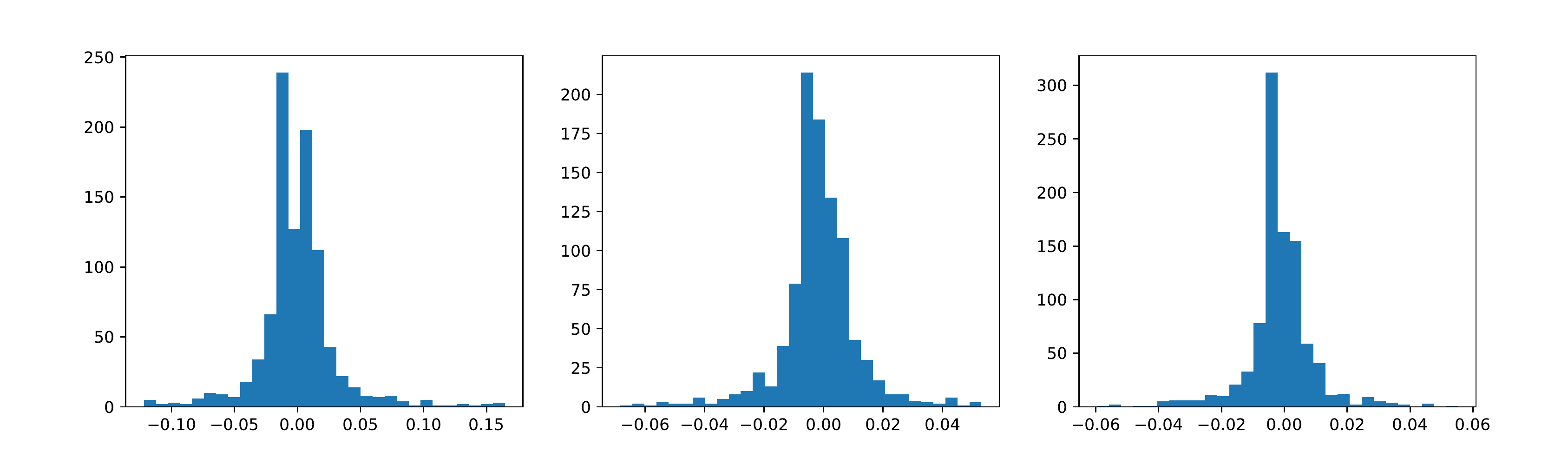}
    \vskip -12pt
    \caption{Each figure shows a histogram of $\theta_j$ for a fixed node $j$ over 1000 independent trials of GCN on Cora. The 3 nodes are randomly selected from the union of the validation set and test set.}
    \label{fig:theta3}
    \vskip -12pt
\end{figure*}

\subsection{Attack Experiments on Benchmark Datasets}
\label{sec:attack-exp}
\vpara{Experiment setup.}
We test attack strategies on 3 popular GNN models, GCN~\citep{kipf2016semi}, GAT~\citep{velivckovic2018graph}, and JK-Net~\citep{xu2018representation}, with 3 public benchmark datasets, Cora, Citeseer, and Pubmed~\citep{sen2008collective}. 

We follow the two-step procedure stated in Section~\ref{sec:setup} to apply the attack strategies. Note that both the proposed strategies and the baseline strategies are designed for the node selection step, and the feature perturbation step (as detailed below) is the same for all methods. 

\emph{Construction of the perturbation vector.} In a real-world scenario, the perturbation vector should be designed according to domain knowledge about the task (e.g., \emph{age} is an important feature that influences the prediction of \emph{income}). For the benchmark datasets, however, we do not know the semantic meaning of the features. So we use proxy models to identify a handful of important features to be perturbed. In particular, for each dataset, we first independently train 20 GCN models as the proxy models of the task. Then we calculate the gradients of the classification loss with respect to the node features, averaged over all the nodes. We treat the top 2\% of features with the largest average gradients as the most important features. And we construct the perturbation vector $\epsilon\in \{-\lambda,0,\lambda\}^D$ ($D$ is the feature dimension and $\lambda>0$ is the perturbation strength) by setting the dimensions corresponding to the (98\%) non-important features as 0, and the dimensions corresponding to the (2\%) important features as $+\lambda$ or $-\lambda$, depending on the sign of the average gradients. The same constant perturbation vector $\epsilon$ is then used for attacking all victim models (GCN, GAT, and JK-Net) in all trials on each dataset.

We note that the GCNs used as proxy models are trained independently of the victim models to be attacked, using different data splits. The construction of $\epsilon$ is therefore completely agnostic of the model information of the victim models. Due to page limit, we leave more details of the experiment setup to Appendix~\ref{appendix:exp}.

\vpara{Experiment results.}
We provide the attack experiment results in Table~\ref{tab:results}. We show the model accuracy after applying each attack strategy in each dataset and model combination, the lower the better. We also include the model accuracy without attack (\textbf{None}) and with an attack under random node selection (\textbf{Random}) for reference. 

As can be seen in Table~\ref{tab:results}, both the proposed attack strategies achieve better attack performance than all baselines on all but 3 setups, out of the 18 setups in total. And most of the differences are statistically significant. We highlight that, compared to the strongest baseline, GC-RWCS, our methods are simpler, and have fewer hyper-parameters and better interpretation. In addition, the neighbor-removal heuristic also contributes to the performance of GC-RWCS method, while our methods outperform GC-RWCS without such additional heuristics.

\subsection{Visualizing the Distributions of $\theta$}
We also empirically investigate the distributions of $\theta$ to see how likely their PDFs are non-increasing on the positive domain. In particular, given the parameters of a well-trained GCN, we are able to approximately calculate $\theta$ with Eq.~(\ref{eq:theta})\footnote{We can only do it approximately because we do not know $\rho$. For the visualization, we just set $\rho=1$.}.
We train a GCN on Cora and get one set of $\theta$. We repeat this process with 1000 independent model initializations and get 1000 sets of $\theta$. Then we can visualize a histogram over the 1000 values of $\theta_j$ for each node $j$. In Figure~\ref{fig:theta3}, we show the histograms of 3 randomly selected nodes. We show the histograms of more randomly selected nodes in Appendix~\ref{appendix:hist}. In most cases, the empirical probability density decreases when $\theta_j>0$, which is the assumption required for the expected mis-classification rate to be submodular in Proposition~\ref{prop:submodular}.

\subsection{Case Study of the Selected Nodes}
\begin{figure}
    \centering
    \includegraphics[width=\textwidth]{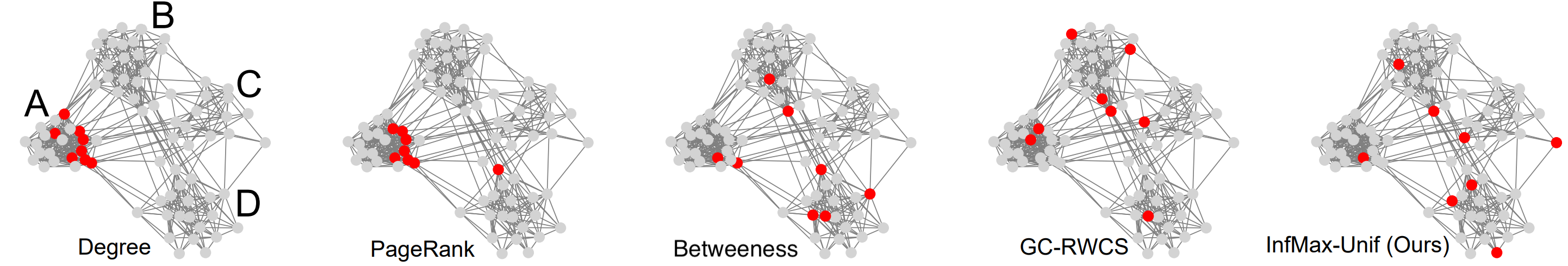}
    \caption{Visualization of the selected nodes by different strategies on a synthetic graph generated by a stochastic blockmodel with 4 communities (A, B, C, and D). The red nodes are the selected nodes.}
    \label{fig:vis-nodes}
\end{figure}

To gain better qualitative understandings of the selected nodes returned by different strategies, we generate a small graph using stochastic blockmodel~\citep{holland1983stochastic} with 4 communities, so that we can visualize the selected nodes. As can be seen in Figure~\ref{fig:vis-nodes}, better performed strategies (GC-RWCS and ours) tend to spread the selected nodes more evenly over the communities. This aligns well with the common belief in the influence maximization literature~\citep{schoenebeck2019think} that a good solution of the selected nodes under linear threshold models tend to have nodes far apart from each other on the graph.

\section{Conclusion}

We present a formal analysis of a restricted and realistic attack setup on Graph Neural Networks (GNNs). By establishing a novel connection between the original attack problem and an influence maximization problem on a linear threshold model, we develop a group of efficient and effective black-box attack strategies with nice interpretations. Extensive empirical results demonstrate the effectiveness of the proposed methods on multiple types of GNNs. 

While the effectiveness of the proposed attack strategies reflects a systematic weakness of existing popular GNN architectures, which could be mis-used for malicious purposes. On the other hand, we believe the interpretable connection between the adversarial attack and the well-known influence maximization problem may motivate more robust GNN designs in future work.

\bibliographystyle{plainnat}
\bibliography{citation}

\begin{thebibliography}{34}
\providecommand{\natexlab}[1]{#1}
\providecommand{\url}[1]{\texttt{#1}}
\expandafter\ifx\csname urlstyle\endcsname\relax
  \providecommand{\doi}[1]{doi: #1}\else
  \providecommand{\doi}{doi: \begingroup \urlstyle{rm}\Url}\fi

\bibitem[Bojchevski and G{\"u}nnemann(2018)]{bojchevski2018adversarial}
Aleksandar Bojchevski and Stephan G{\"u}nnemann.
\newblock Adversarial attacks on node embeddings via graph poisoning.
\newblock \emph{arXiv preprint arXiv:1809.01093}, 2018.

\bibitem[Chang et~al.(2020)Chang, Rong, Xu, Huang, Zhang, Cui, Zhu, and
  Huang]{chang2020restricted}
Heng Chang, Yu~Rong, Tingyang Xu, Wenbing Huang, Honglei Zhang, Peng Cui, Wenwu
  Zhu, and Junzhou Huang.
\newblock A restricted black-box adversarial framework towards attacking graph
  embedding models.
\newblock In \emph{AAAI Conference on Artificial Intelligence}, 2020.

\bibitem[Chen et~al.(2017)Chen, Zhang, Sharma, Yi, and Hsieh]{chen2017zoo}
Pin-Yu Chen, Huan Zhang, Yash Sharma, Jinfeng Yi, and Cho-Jui Hsieh.
\newblock Zoo: Zeroth order optimization based black-box attacks to deep neural
  networks without training substitute models.
\newblock In \emph{Proceedings of the 10th ACM workshop on artificial
  intelligence and security}, pages 15--26, 2017.

\bibitem[Choromanska et~al.(2015)Choromanska, Henaff, Mathieu, Arous, and
  LeCun]{choromanska2015loss}
Anna Choromanska, Mikael Henaff, Michael Mathieu, G{\'e}rard~Ben Arous, and
  Yann LeCun.
\newblock The loss surfaces of multilayer networks.
\newblock In \emph{Artificial intelligence and statistics}, pages 192--204,
  2015.

\bibitem[Dai et~al.(2018)Dai, Li, Tian, Huang, Wang, Zhu, and
  Song]{dai2018adversarial}
Hanjun Dai, Hui Li, Tian Tian, Xin Huang, Lin Wang, Jun Zhu, and Le~Song.
\newblock Adversarial attack on graph structured data.
\newblock \emph{arXiv preprint arXiv:1806.02371}, 2018.

\bibitem[Entezari et~al.(2020)Entezari, Al-Sayouri, Darvishzadeh, and
  Papalexakis]{entezari2020all}
Negin Entezari, Saba~A Al-Sayouri, Amirali Darvishzadeh, and Evangelos~E
  Papalexakis.
\newblock All you need is low (rank) defending against adversarial attacks on
  graphs.
\newblock In \emph{Proceedings of the 13th International Conference on Web
  Search and Data Mining}, pages 169--177, 2020.

\bibitem[Fan et~al.(2019)Fan, Ma, Li, He, Zhao, Tang, and Yin]{fan2019graph}
Wenqi Fan, Yao Ma, Qing Li, Yuan He, Eric Zhao, Jiliang Tang, and Dawei Yin.
\newblock Graph neural networks for social recommendation.
\newblock In \emph{The World Wide Web Conference}, pages 417--426, 2019.

\bibitem[Granovetter(1978)]{granovetter1978threshold}
Mark Granovetter.
\newblock Threshold models of collective behavior.
\newblock \emph{American journal of sociology}, 83\penalty0 (6):\penalty0
  1420--1443, 1978.

\bibitem[Hamilton et~al.(2017)Hamilton, Ying, and
  Leskovec]{hamilton2017inductive}
Will Hamilton, Zhitao Ying, and Jure Leskovec.
\newblock Inductive representation learning on large graphs.
\newblock In \emph{Advances in neural information processing systems}, pages
  1024--1034, 2017.

\bibitem[Holland et~al.(1983)Holland, Laskey, and
  Leinhardt]{holland1983stochastic}
Paul~W Holland, Kathryn~Blackmond Laskey, and Samuel Leinhardt.
\newblock Stochastic blockmodels: First steps.
\newblock \emph{Social networks}, 5\penalty0 (2):\penalty0 109--137, 1983.

\bibitem[Iyer and Bilmes(2015)]{iyer2015submodular}
Rishabh Iyer and Jeffrey Bilmes.
\newblock Submodular point processes with applications to machine learning.
\newblock In \emph{Artificial Intelligence and Statistics}, pages 388--397,
  2015.

\bibitem[Jin et~al.(2020)Jin, Li, Xu, Wang, and Tang]{jin2020adversarial}
Wei Jin, Yaxin Li, Han Xu, Yiqi Wang, and Jiliang Tang.
\newblock Adversarial attacks and defenses on graphs: A review and empirical
  study.
\newblock \emph{arXiv preprint arXiv:2003.00653}, 2020.

\bibitem[Kawaguchi(2016)]{kawaguchi2016deep}
Kenji Kawaguchi.
\newblock Deep learning without poor local minima.
\newblock In \emph{Advances in neural information processing systems}, pages
  586--594, 2016.

\bibitem[Kempe et~al.(2003)Kempe, Kleinberg, and Tardos]{kempe2003maximizing}
David Kempe, Jon Kleinberg, and {\'E}va Tardos.
\newblock Maximizing the spread of influence through a social network.
\newblock In \emph{Proceedings of the ninth ACM SIGKDD international conference
  on Knowledge discovery and data mining}, pages 137--146, 2003.

\bibitem[Kipf and Welling(2016)]{kipf2016semi}
Thomas~N Kipf and Max Welling.
\newblock Semi-supervised classification with graph convolutional networks.
\newblock \emph{arXiv preprint arXiv:1609.02907}, 2016.

\bibitem[Li et~al.(2017)Li, Ma, Guo, and Mei]{li2017deepcas}
Cheng Li, Jiaqi Ma, Xiaoxiao Guo, and Qiaozhu Mei.
\newblock Deepcas: An end-to-end predictor of information cascades.
\newblock In \emph{Proceedings of the 26th international conference on World
  Wide Web}, pages 577--586, 2017.

\bibitem[Li et~al.(2020)Li, Guo, and Chen]{li2020practical}
Qizhang Li, Yiwen Guo, and Hao Chen.
\newblock Practical no-box adversarial attacks against dnns.
\newblock \emph{Advances in Neural Information Processing Systems}, 33, 2020.

\bibitem[Lin and Bilmes(2011)]{lin2011class}
Hui Lin and Jeff Bilmes.
\newblock A class of submodular functions for document summarization.
\newblock In \emph{Proceedings of the 49th Annual Meeting of the Association
  for Computational Linguistics: Human Language Technologies}, pages 510--520,
  2011.

\bibitem[Ma et~al.(2020)Ma, Ding, and Mei]{ma2020black}
Jiaqi Ma, Shuangrui Ding, and Qiaozhu Mei.
\newblock Towards more practical adversarial attacks on graph neural networks.
\newblock \emph{Advances in neural information processing systems}, 2020.

\bibitem[Mossel and Roch(2007)]{mossel2007submodularity}
Elchanan Mossel and Sebastien Roch.
\newblock On the submodularity of influence in social networks.
\newblock In \emph{Proceedings of the thirty-ninth annual ACM symposium on
  Theory of computing}, pages 128--134, 2007.

\bibitem[Schoenebeck et~al.(2019)Schoenebeck, Tao, and
  Yu]{schoenebeck2019think}
Grant Schoenebeck, Biaoshuai Tao, and Fang-Yi Yu.
\newblock Think globally, act locally: On the optimal seeding for nonsubmodular
  influence maximization.
\newblock \emph{Approximation, Randomization, and Combinatorial Optimization.
  Algorithms and Techniques (APPROX/RANDOM 2019)}, 145:\penalty0 39, 2019.

\bibitem[Sen et~al.(2008)Sen, Namata, Bilgic, Getoor, Galligher, and
  Eliassi-Rad]{sen2008collective}
Prithviraj Sen, Galileo Namata, Mustafa Bilgic, Lise Getoor, Brian Galligher,
  and Tina Eliassi-Rad.
\newblock Collective classification in network data.
\newblock \emph{AI magazine}, 29\penalty0 (3):\penalty0 93--93, 2008.

\bibitem[Sun et~al.(2018)Sun, Dou, Yang, Wang, Yu, and Li]{sun2018adversarial}
Lichao Sun, Yingtong Dou, Carl Yang, Ji~Wang, Philip~S Yu, and Bo~Li.
\newblock Adversarial attack and defense on graph data: A survey.
\newblock \emph{arXiv preprint arXiv:1812.10528}, 2018.

\bibitem[Sun et~al.(2019)Sun, Wang, Tang, Hsieh, and Honavar]{sun2019node}
Yiwei Sun, Suhang Wang, Xianfeng Tang, Tsung-Yu Hsieh, and Vasant Honavar.
\newblock Node injection attacks on graphs via reinforcement learning.
\newblock \emph{arXiv preprint arXiv:1909.06543}, 2019.

\bibitem[Tang et~al.(2020)Tang, Ma, Chen, Guo, Wang, Zeng, and
  Zhan]{tang2020adversarial}
Haoteng Tang, Guixiang Ma, Yurong Chen, Lei Guo, Wei Wang, Bo~Zeng, and Liang
  Zhan.
\newblock Adversarial attack on hierarchical graph pooling neural networks.
\newblock \emph{arXiv preprint arXiv:2005.11560}, 2020.

\bibitem[Veli{\v{c}}kovi{\'c} et~al.(2018)Veli{\v{c}}kovi{\'c}, Cucurull,
  Casanova, Romero, Li{\`o}, and Bengio]{velivckovic2018graph}
Petar Veli{\v{c}}kovi{\'c}, Guillem Cucurull, Arantxa Casanova, Adriana Romero,
  Pietro Li{\`o}, and Yoshua Bengio.
\newblock Graph attention networks.
\newblock In \emph{International Conference on Learning Representations}, 2018.

\bibitem[Wu et~al.(2019)Wu, Wang, Tyshetskiy, Docherty, Lu, and
  Zhu]{wu2019adversarial}
Huijun Wu, Chen Wang, Yuriy Tyshetskiy, Andrew Docherty, Kai Lu, and Liming
  Zhu.
\newblock Adversarial examples for graph data: deep insights into attack and
  defense.
\newblock In \emph{Proceedings of the 28th International Joint Conference on
  Artificial Intelligence}, pages 4816--4823. AAAI Press, 2019.

\bibitem[Wu et~al.(2020)Wu, Pan, Chen, Long, Zhang, and
  Philip]{wu2020comprehensive}
Zonghan Wu, Shirui Pan, Fengwen Chen, Guodong Long, Chengqi Zhang, and S~Yu
  Philip.
\newblock A comprehensive survey on graph neural networks.
\newblock \emph{IEEE Transactions on Neural Networks and Learning Systems},
  2020.

\bibitem[Xu et~al.(2019)Xu, Chen, Liu, Chen, Weng, Hong, and
  Lin]{xu2019topology}
Kaidi Xu, Hongge Chen, Sijia Liu, Pin-Yu Chen, Tsui-Wei Weng, Mingyi Hong, and
  Xue Lin.
\newblock Topology attack and defense for graph neural networks: An
  optimization perspective.
\newblock \emph{arXiv preprint arXiv:1906.04214}, 2019.

\bibitem[Xu et~al.(2018)Xu, Li, Tian, Sonobe, Kawarabayashi, and
  Jegelka]{xu2018representation}
Keyulu Xu, Chengtao Li, Yonglong Tian, Tomohiro Sonobe, Ken-ichi Kawarabayashi,
  and Stefanie Jegelka.
\newblock Representation learning on graphs with jumping knowledge networks.
\newblock \emph{arXiv preprint arXiv:1806.03536}, 2018.

\bibitem[Ying et~al.(2018)Ying, He, Chen, Eksombatchai, Hamilton, and
  Leskovec]{ying2018graph}
Rex Ying, Ruining He, Kaifeng Chen, Pong Eksombatchai, William~L Hamilton, and
  Jure Leskovec.
\newblock Graph convolutional neural networks for web-scale recommender
  systems.
\newblock In \emph{Proceedings of the 24th ACM SIGKDD International Conference
  on Knowledge Discovery \& Data Mining}, pages 974--983, 2018.

\bibitem[Yu et~al.(2017)Yu, Yin, and Zhu]{yu2017spatio}
Bing Yu, Haoteng Yin, and Zhanxing Zhu.
\newblock Spatio-temporal graph convolutional networks: A deep learning
  framework for traffic forecasting.
\newblock \emph{arXiv preprint arXiv:1709.04875}, 2017.

\bibitem[Z{\"u}gner and G{\"u}nnemann(2019)]{zugner_adversarial_2019}
Daniel Z{\"u}gner and Stephan G{\"u}nnemann.
\newblock Adversarial attacks on graph neural networks via meta learning.
\newblock In \emph{International Conference on Learning Representations
  (ICLR)}, 2019.

\bibitem[Z{\"u}gner et~al.(2018)Z{\"u}gner, Akbarnejad, and
  G{\"u}nnemann]{zugner2018adversarial}
Daniel Z{\"u}gner, Amir Akbarnejad, and Stephan G{\"u}nnemann.
\newblock Adversarial attacks on neural networks for graph data.
\newblock In \emph{Proceedings of the 24th ACM SIGKDD International Conference
  on Knowledge Discovery \& Data Mining}, pages 2847--2856, 2018.

\end{thebibliography}

\newpage
\appendix
\section{Proofs}
We first give a more precise and restated version (Assumption~\ref{assum:relu-re}) of Assumption~\ref{assum:relu}, and introduce Lemma~\ref{lemma:xu} about GCN, which is proved by \citet{xu2018representation}.

\begin{assumption} [\citet{xu2018representation} Restated.]
\label{assum:relu-re}
Recall that a ReLU function can be written as
\[\text{ReLU}(x) = x\cdot \ind{x>0}.\]
Suppose there are $R$ ReLU functions in the GCN model and we index them with $i=1,2,\cdots, R$. This assumption assumes that the $i$-th ReLU functions, for $i=1,2,\cdots, R$, is replaced by the following function,
\[\text{ReLU}_i(x) = x\cdot z_i,\]
where $z_1,z_2,\cdots, z_R \distas{i.i.d.} \text{Bernoulli}(\gamma)$.

This assumption implies that all paths in the computation graph of an $L$-layer GCN model are independently activated with the same probability $\rho=\gamma^L$.
\end{assumption}

\begin{lemma} [\citet{xu2018representation}.]
\label{lemma:xu}
Given an $L$-layer GCN, under Assumption~\ref{assum:relu}, for any node $i,j\in V$, 
\begin{equation}
     \E_{\text{path}}\left[\frac{\partial H_j}{\partial X_i}\right]=\rho [M^L]_{ji} \cdot \left(\prod_{l=L}^{1} W^{(l)}\right),
      \label{randomwalk}
\end{equation}
where $M\in \reals^{N\times N}$ is the random walk transition matrix, i.e., for any $1\le i,j\le N$, $M_{ij} = 1/|\mathcal{N}_i|$ if $(i, j)\in E$ or $i=j$, and $M_{ij} = 0$ otherwise.
\end{lemma}

\subsection{Proof for Proposition~\ref{prop:opt_ltm}}
\label{appendix:proof-opt_ltm}
\begin{proof}
Recall that $\bH(S) = \E_{\text{path}}[H(S)] = \E_{\text{path}}[f(X(S))]$. We first show $\bH(S)$ is a linear function of $X(S)$, which suffices to show that, for any $i\in V$ and $1\le l \le L$, $\E_{\text{path}}[H^{(l)}_i(S)]$ is a linear function of $\E_{\text{path}}[H^{(l-1)}(S)]$. When $l=L$,
\[
\E_{\text{path}}[H^{(L)}_i(S)] = \sum_{j\in \mathcal{N}_i} \alpha_{ij} W^{(L)} \E_{\text{path}}[H_j^{(L-1)}(S)],
\]
so the statement holds. When $1\le l < L$, under Assumption~\ref{assum:relu}, suppose each ReLU activates independently with probability $p$.
\begin{align*}
    \E_{\text{path}}H_i^{(l)} &= \E_{\text{path}}\left[\sigma\sbra \sum_{j\in \mathcal{N}_i} \alpha_{ij} W^{(l)} H_j^{(l-1)} \sket\right] \\
    &= p \sum_{j\in \mathcal{N}_i} \alpha_{ij} W^{(L)} \E_{\text{path}}[H_j^{(l-1)}(S)],
\end{align*}
so the statement also holds. Therefore $\bH(S)$ is a linear function of $X(S)$. In particular, $\E_{\text{path}}[H] = \bH(\emptyset)$ is a linear function of $X$.

We know that $X_i(S) = X_i + \epsilon$ for $i\in S$ and $X_i(S) = X_i$ for $i\notin S$. And by Lemma~\ref{lemma:xu}, we can rewrite $\bH(S)$ in terms of $\bH(\emptyset)$ and $\epsilon$. For any $j\in V$,
\[
    \bH_j(S) = \bH_j(\emptyset) + \sum_{i\in S}\rho [M^L]_{ji} \cdot \left(\prod_{l=L}^{1} W^{(l)}\right)^T\epsilon.
\]
In Section~\ref{sec:connection}, we have defined $B=M^L$ and $W = \rho\prod_{l=L}^{1} W^{(l)}$, so \begin{equation}
    \label{eq:linear}
    \bH_j(S) = \bH_j(\emptyset) + W^T\epsilon \sum_{i\in S}B_{ji}.
\end{equation}

Now we look at the objective~(\ref{eq:opt_original}). If we replace $H(S)$ with $\bH(S)$ in this objective and plug Eq.~(\ref{eq:linear}) into it, then for each $j\in V$, we have
\begin{align}
    & \ind{\max_{k\in \{1,\cdots, K\}} \bH_{jk}(S)\neq \bH_{jy_j}(S)} \nonumber\\
    =& \ind{\bH_{j\hk_j}(S) > \bH_{jy_j}(S)} \nonumber\\
    =& \ind{\bH_{j\hk_j}(\emptyset) + W_{\hk_j}^T\epsilon\cdot \sum_{i\in S}B_{ji} > \bH_{jy_j}(\emptyset) + W_{y_j}^T\epsilon\cdot \sum_{i\in S}B_{ji}} \nonumber\\
    =& \ind{\sum_{i\in S}B_{ji} > \frac{\bH_{jy_j}(\emptyset) - \bH_{j\hk_j}(\emptyset)}{(W_{\hk_j} - W_{y_j})^T\epsilon}} \nonumber\\
    =& \ind{\sum_{i\in S}B_{ji} > \theta_j}, \nonumber
\end{align}
where we have defined $\hk_j=\argmax_{k=1,\cdots,K}\bH_{jk}(S)$ and recall the definition of $\theta_j$ in Eq.~(\ref{eq:theta}). 

Therefore we get the optimization problem~(\ref{eq:opt_ltm})
\begin{align}
    \max_{S\in C_{r,m}}\quad&\sum_{j=1}^N\ind{\sum_{i\in S}B_{ji} > \theta_j}. \nonumber
\end{align}
\end{proof}

\subsection{Proof for Lemma~\ref{lemma:np}} 
\label{appendix:proof-np}
The proof follows similarly as the proof of Theorem 2.4 in \citet{kempe2003maximizing}.
\begin{proof}
We prove by reducing the NP-complete Set Cover problem to the influence maximization problem on directed biparatite graph with a linear threshold model. The Set Cover problem is defined as following. Suppose we have a ground set $U = \{u_1, u_2, \cdots, u_n\}$ and a group of $m$ subsets of $U$, $S_1, S_2, \cdots, S_m$. The goal is to determine whether there exists $r$ ( $r < n$ and $r < m$) of the subsets whose union equals to $U$. 

For any instance of the Set Cover problem, we can construct a bipartite graph with the first side having $m$ nodes (each one corresponding to a given subset of $U$), and the second side having $n$ nodes (each one corresponding to an element of $U$). There are only links going from the the first side to the second side. There will be a link with constant influence score $\alpha > 0$ from a node on the first side to the second side if and only if the corresponding subset contains that element in $U$. Finally the node-specific thresholds of each node on the second side is set as $\alpha/2$. And the influence maximization problem asks to select $r$ nodes on the graph to maximize the number of activated nodes. The Set Cover problem is then solved by deciding if the maximized number of activated nodes on the bipartite graph is greater than $n+r$. 
\end{proof}

\subsection{Proof for Proposition~\ref{prop:submodular}}
\label{appendix:proof-submodular}
\begin{proof}
We first show that the expected mis-classifcation rate $h(S)$ can be written in terms of the marginal CDFs of $\theta$. 

\begin{align*}
    h(S) &= \E_{\theta_1,\cdots, \theta_N}\sum_{j=1}^N\ind{\sum_{i\in S}B_{ji} > \theta_j}\\
    &= \sum_{j=1}^N \E_{\theta_1,\cdots, \theta_N} \ind{\sum_{i\in S}B_{ji} > \theta_j} \\
    &= \sum_{j=1}^N \E_{\theta_j} \ind{\sum_{i\in S}B_{ji} > \theta_j} \\
    &= \sum_{j=1}^N P_j\left(\sum_{i\in S}B_{ji} > \theta_j\right) \\
    &= \sum_{j=1}^N F_j\left(\sum_{i\in S}B_{ji}\right),
\end{align*}
where $P_j$ is the marginal probability of $\theta_j$.

Since $B_{ji} \ge 0$, so $\sum_{i\in S}B_{ji}$ is a non-decreasing submodular function of $S$ with a lower bound 0. Each CDF $F_j$ is non-decreasing by definition, if it is also individually concave at the domain $[0, +\infty)$, we know $F_j\left(\sum_{i\in S}B_{ji}\right)$ is submodular w.r.t. $S$ and hence $h(S)$ is submodular.

\end{proof}

\begin{figure*}
    \centering
    \includegraphics[width=0.95\textwidth]{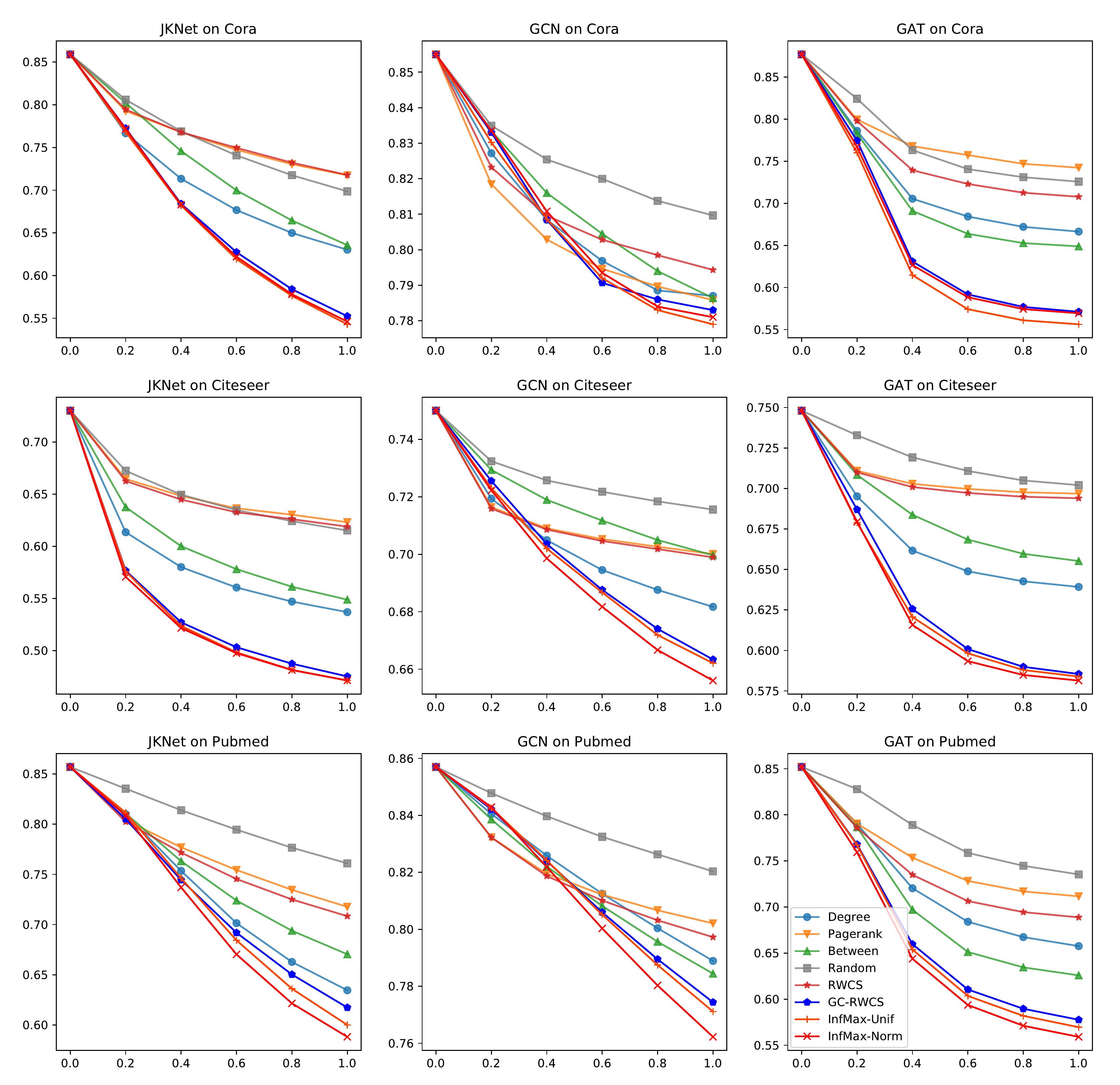}
    \caption{The attack performances with varying perturbation strengths (from 0 to 1). Each figure corresponds to a dataset-model combination. The x-axis indicates the value of $\lambda$ and the y-axis indicates the classification accuracy after attack. The threshold is set as 10\% and all other experiment setups are the same as those in Table~\ref{tab:old-results}.} 
    \label{fig:lambda}
\end{figure*}

\section{More Experiment Details}
\subsection{Baseline Methods}
\label{appendix:baseline}
\vpara{Definitions of the node centralities.} For each node $i$, the Degree centrality score is defined as $C_D(i) \triangleq \frac{|\mathcal{N}_i|}{N}$; the Betweenness centrality score is defined as $C_B(i)\triangleq \sum_{j\neq i, k\neq i, j < k} \frac{g_{jk}(i)}{g_{jk}}$, where $g_{jk}$ is the number of shortest paths connecting node $j$ and $k$ and $g_{jk}(i)$ is the number of shortest paths that node $i$ is on; the PageRank centrality score is defined as the stationary scores achieved by iteratively updating $PR(i) = \frac{1-\alpha}{N} + \alpha\sum_{j\in \mathcal{N}_i}\frac{PR(j)}{|\mathcal{N}_j|}$ and we set the hyper-parameter $\alpha=0.85$. 

\vpara{Detailed descriptions of GC-RWCS.} GC-RWCS further applies a few heuristics on top of RWCS to achieve better mis-classification rate. Specifically, it iteratively selects nodes one by one up to $r$ nodes, based on a dynamic importance score, i.e., $I_t(i)=\sum_{j=1}^N [Q_t]_{ji}$ for the $t$-th iteration. $Q_t\in \{0, 1\}^{N\times N}$ is a binary matrix that is dynamically updated over $t$. At the initial iteration, $Q_1$ is obtained by binarizing $M^L$, assigning  $1$ to the top $l$ nonzero entries in each row of $M^L$ and $0$ to other entries. For $t>1$, suppose the node $i$ is selected at the $t-1$ iteration, then $Q_t$ is obtained from $Q_{t-1}$ by setting to zero for all the rows where the elements of the $i$-th column is 1 in $Q_{t-1}$. GC-RWCS also applies another heuristic that, after each iteration, remove the $k$-hop neighbors of the selected node from the candidate set in the subsequent iterations. In the experiment, we set the hyper-parameters of GC-RWCS $L=4$, $l=30$, and $k=1$ as suggested in their original paper. The iterative-selection process in GC-RWCS (without removing the $k$-hop neighbors) gives equivalent results as InfMax-Unif if we replace the matrix $B$ in InfMax-Unif by $Q_1$ and set $a=1$. 

\subsection{More Details for the Experiment Setup}
\label{appendix:exp}
% TODO: state the detailed new experiment setup as well as the original setup
We closely follow the experiment setup by \citet{ma2020black}. We set the number of layers of GCN and GAT as 2, and that of JK-Net as 7. We use the implementations of all models in Deep Graph Library\footnote{Website: \url{https://www.dgl.ai}. Apache license: \url{https://github.com/dmlc/dgl/blob/master/LICENSE}.}. We randomly split each dataset by 60\%, 20\% and 20\% as the training, validation, and test sets and run 40 independent trials for each model and dataset combination. We apply the attack strategies following the two-step procedure stated in Section~\ref{sec:setup}. For the node selection step, we limit the number of nodes to be attacked, $r$, as 1\% of the graph size for each dataset. We test on two setups of the node degree threshold, $m$, by setting it equal to the lowest degree of the top 10\% and 30\% nodes respectively. 

For the feature perturbation step, we follow a similar way as in \citet{ma2020black} to construct the constant perturbation vector $\epsilon$. But our construction is more strictly black-box compared to \citet{ma2020black}. 
Specifically, we first train 20 GCN models on each dataset. Then for each trained GCN (indexed by $k=1,\cdots,20$), we calculate a set of (signed) importance score $\bar{g}^{(k)}_j=\frac{1}{N}\sum_{i=1}^N\frac{\partial \mathcal{L}(H,y)}{\partial X_{i,j}}$ for each dimension of feature, $j=1,\cdots, D$, and $\mathcal{L}(\cdot, \cdot)$ is the classification loss. Intuitively, $\bar{g}^{(k)}_j$'s are the average gradients over all nodes and provides an indicator of how important an feature is to the classification task at a coarse population level. We then select the top 2\% important features by the following procedure: 1) we first filter the features $j$'s to make sure that at least 80\% of $\{\bar{g}^{(k)}_j\}_{k=1}^{20}$ have the same sign; 2) then we select the top $0.02 D$ features with largest average scores, $\frac{1}{20}\sum_{k=1}^{20}|\bar{g}^{(k)}_j|$. Then we construct the perturbation vector $\epsilon\in \{-\lambda,0,\lambda\}^D$ ($\lambda$ is the perturbation strength and is set as 10 in our experiments) by setting the dimensions corresponding to the (98\%) non-important features as 0, and the dimensions corresponding to the (2\%) important features as $+\lambda$ or $-\lambda$, with the sign determined by the majority of $\{\bar{g}^{(k)}_j\}_{k=1}^{20}$. The same constant perturbation vectors are then added to all models in all trials of the experiment. We highlight that \textbf{the 20 GCN models used to construct the perturbation vector are independently trained with different training splits compared to the models (GCN, GAT, and JK-Net) we attack in Table~\ref{tab:results}}. This makes our experiment setup completely black-box. Notably, when attacking GAT and JK-Net, the construction of perturbation is even unaware of the exact GNN architectures of the victim models.

\begin{figure*}
    \centering
    \includegraphics[width=\textwidth]{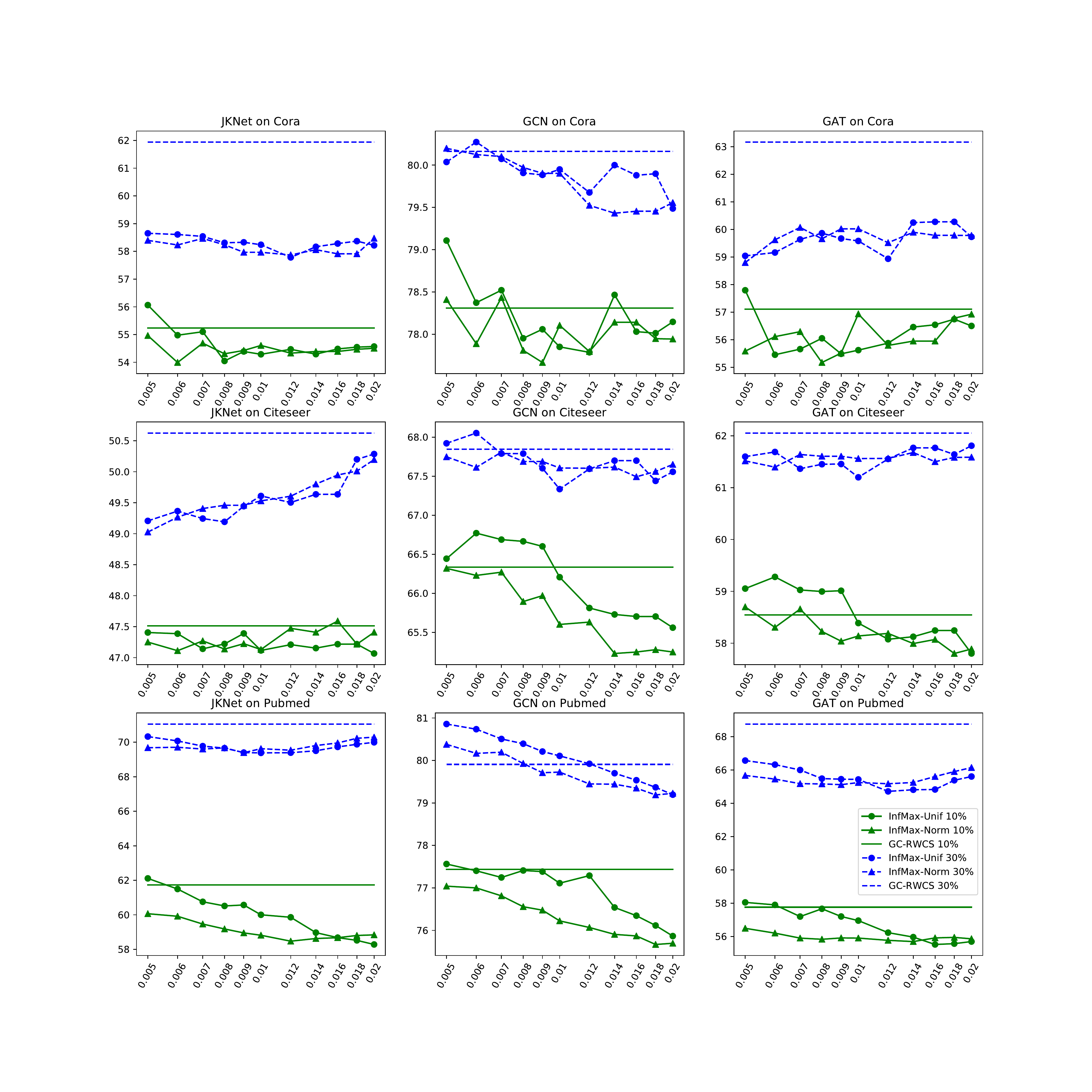}
    \caption{Sensitivity analysis of the hyper-parameters $a$ and $\sigma$. The experiment setups are the same as those in Section~\ref{sec:attack-exp}. Each figure corresponds to a dataset-model combination. The x-axis indicates the value of $a$ or $\sigma$ while the y-axis indicates the classification accuracy after attack. The results under the threshold 10\% are plotted in green while the results under the threshold 30\% are plotted in blue. In addition to the proposed InfMax-Unif and InfMax-Norm, we also plot the results of GC-RWCS as the constant dashed lines for references. The plots are made in log-scale for the x-axis.}
    \label{fig:sensitivity}
\end{figure*}

\section{Additional Experiments}
\label{appendix:additional}
\subsection{Results under the Experiment Setup by \citet{ma2020black}}
In a previous version of this paper, we conducted experiments under a setup exactly the same as that by \citet{ma2020black}. The main difference between our current setup (as detailed in Appendix~\ref{appendix:exp}) and their setup is that we now use GCN models independent of the victim models to construct the constant perturbation vector, while they directly use the victim models to construct the perturbation vector. Although the use of the model information is extremely limited in the setup by \citet{ma2020black}, it is not completely black-box. Nevertheless, in this section, we provide the experiment results under the setup by \citet{ma2020black} for easier comparison with existing literature.

As can be seen in Table~\ref{tab:old-results}, the overall trend is almost the same as Table~\ref{tab:results}.

\begin{table*}
\caption{Experiment results under the setup by \citet{ma2020black}. Summary of the attack performance in terms of test accuracy (\%), the lower the better attack. \textbf{Bold} denotes the best performing strategy in each setup. \uline{Underline} indicates our strategy outperforms all the baseline strategies. Asterisk (*) means the difference between our strategy and the best baseline strategy is statistically significant by a pairwise t-test at significance level 0.05. The error bar ($\pm$) denotes the standard error of the mean by 40 independent trials. We test on two setups of the node degree threshold, $m$, by setting it equal to the lowest degree of the top 10\% and 30\% nodes respectively.}
\label{tab:old-results}
\resizebox{\textwidth}{!}{%
\begin{tabular}{l|lll|lll|lll}
\toprule
                       & \multicolumn{3}{c|}{Cora}                                                                   & \multicolumn{3}{c|}{Citeseer}                                                               & \multicolumn{3}{c}{Pubmed}                                                                 \\
% \midrule
Method                 & JKNet                 & GCN                          & GAT                          & JKNet                 & GCN                          & GAT                          & JKNet                 & GCN                          & GAT                          \\
\midrule
None                   & $85.9\pm0.1$                 & $85.5\pm0.2$                 & $87.7\pm0.2$                 & $73.0\pm0.2$                 & $75.0\pm0.2$                 & $74.8\pm0.2$                 & $85.7\pm0.1$                 & $85.7\pm0.1$    & $85.2\pm0.1$                 \\
\midrule
\multicolumn{10}{c}{Threshold 10\%}                                                                                                                                                                                                                                                                           \\
\midrule
Random                 & $69.9\pm1.1$                 & $81.7\pm0.3$                 & $72.6\pm0.6$                 & $61.5\pm0.9$                 & $71.6\pm0.2$                 & $70.2\pm0.5$                 & $76.1\pm0.6$                 & $82.0\pm0.3$                 & $73.5\pm0.3$                 \\
Degree                 & $63.0\pm1.4$                 & $78.7\pm0.4$                 & $66.6\pm0.7$                 & $53.7\pm0.9$                 & $68.2\pm0.3$                 & $63.9\pm0.5$                 & $63.5\pm0.9$                 & $78.9\pm0.5$                 & $65.8\pm0.7$                 \\
Pagerank               & $71.7\pm0.9$                 & $80.1\pm0.3$                 & $74.2\pm0.5$                 & $62.3\pm0.6$                 & $70.0\pm0.3$                 & $69.7\pm0.3$                 & $71.8\pm0.8$                 & $80.2\pm0.3$                 & $71.2\pm0.3$                 \\
Betweenness            & $63.6\pm1.4$                 & $80.2\pm0.4$                 & $64.9\pm0.5$                 & $54.9\pm1.0$                 & $70.0\pm0.3$                 & $65.5\pm0.5$                 & $67.0\pm1.0$                 & $78.4\pm0.5$                 & $62.6\pm0.6$                 \\
RWCS                   & $71.8\pm0.8$                 & $80.3\pm0.4$                 & $70.8\pm0.5$                 & $61.9\pm0.6$                 & $69.9\pm0.3$                 & $69.4\pm0.3$                 & $70.8\pm0.8$                 & $79.7\pm0.3$                 & $68.9\pm0.4$                 \\
GC-RWCS                & $55.2\pm1.5$                 & $78.3\pm0.5$                 & $57.1\pm0.6$                 & $47.5\pm1.0$                 & $66.3\pm0.5$                 & $58.5\pm0.6$                 & $61.7\pm1.1$                 & $77.4\pm0.6$                 & $57.8\pm0.8$                 \\
InfMax-Unif          & {\ul \boldmath{$54.3\pm1.5$*}}           & {\ul \boldmath{$77.9\pm0.5*$}} & {\ul \boldmath{$55.6\pm0.6*$}} & {\ul \boldmath{$47.1\pm1.0*$}} & {\ul $66.2\pm0.5$}           & {\ul $58.4\pm0.6$}           & {\ul $60.0\pm1.2$*}           & {\ul $77.1\pm0.7$*}           & {\ul $57.0\pm0.9$*}           \\
InfMax-Norm              & {\ul $54.6\pm1.5$*}           & {\ul $78.1\pm0.5$}           & {\ul $56.9\pm0.6$}           & {\ul \boldmath{$47.1\pm1.0*$}} & {\ul \boldmath{$65.6\pm0.5*$}} & {\ul \boldmath{$58.1\pm0.6*$}} & {\ul \boldmath{$58.8\pm1.1*$}}           & {\ul \boldmath{$76.2\pm0.7*$}}           & {\ul \boldmath{$55.9\pm1.0*$}}           \\
\midrule
\multicolumn{10}{c}{Threshold 30\%}                                                                                                                                                                                                                                                                           \\
\midrule
Random                 & $71.5\pm1.1$                 & $82.1\pm0.3$                 & $74.1\pm0.6$                 & $64.0\pm0.8$                 & $72.4\pm0.2$                 & $71.7\pm0.3$                 & $78.0\pm0.4$                 & $82.4\pm0.3$                 & $76.0\pm0.3$                 \\
Degree                 & $67.5\pm1.2$                 & $81.0\pm0.4$                 & $70.4\pm0.6$                 & $58.4\pm1.0$                 & $70.5\pm0.3$                 & $67.7\pm0.4$                 & $73.2\pm0.8$                 & $81.1\pm0.4$                 & $71.0\pm0.4$                 \\
Pagerank               & $79.4\pm0.5$                 & $82.5\pm0.3$                 & $82.3\pm0.3$                 & $70.2\pm0.3$                 & $72.7\pm0.2$                 & $73.8\pm0.2$                 & $79.9\pm0.3$                 & $82.6\pm0.2$                 & $79.0\pm0.2$                 \\
Betweenness            & $66.9\pm1.3$                 & $81.4\pm0.3$                 & $67.5\pm0.5$                 & $57.7\pm1.0$                 & $70.8\pm0.3$                 & $67.8\pm0.5$                 & $75.3\pm0.5$                 & $80.9\pm0.4$                 & $71.7\pm0.4$                 \\
RWCS                   & $79.2\pm0.5$                 & $82.5\pm0.3$                 & $82.3\pm0.3$                 & $69.9\pm0.3$                 & $72.7\pm0.2$                 & $73.7\pm0.2$                 & $78.2\pm0.3$                 & $81.7\pm0.3$                 & $77.8\pm0.2$                 \\
GC-RWCS                & $61.9\pm1.5$                 & $80.2\pm0.4$                 & $63.2\pm0.5$                 & $50.6\pm1.1$                 & $67.8\pm0.4$                 & $62.1\pm0.6$                 & $71.1\pm0.8$                 & $79.9\pm0.5$                 & $68.8\pm0.4$                 \\
InfMax-Unif          & {\ul $58.2\pm1.5$*}           & {\ul \boldmath{$79.9\pm0.4$}}           & {\ul \boldmath{$59.6\pm0.5*$}} & {\ul $49.6\pm1.0*$}           & {\ul \boldmath{$67.3\pm0.5*$}} & {\ul \boldmath{$61.2\pm0.6*$}} & {\ul \boldmath{$69.4\pm1.0*$}} & $80.1\pm0.5$                 & {\ul $65.4\pm0.5$*}           \\
InfMax-Norm              & {\ul \boldmath{$58.0\pm1.5*$}}           & {\ul $79.9\pm0.5$}           & {\ul $60.0\pm0.5*$}           & {\ul \boldmath{$49.5\pm1.0*$}}           & {\ul $67.6\pm0.5$}           & {\ul $61.6\pm0.6*$}           & {\ul $69.6\pm1.0*$}           & {\ul \boldmath{$79.7\pm0.5$}}           & {\ul \boldmath{$65.2\pm0.5$*} }          \\
\bottomrule
\end{tabular}%
}
\end{table*}

\subsection{Sensitivity Analyses}
\label{appendix:sensitivity}
\vpara{Attack performance with varying perturbation strengths.} In Figure~\ref{fig:lambda}, we demonstrate the attack performances of different attack strategies with varying perturbation strengths. We first observe that the proposed attack strategies with the fixed hyper-parameters ($a=0.01$ for InfMax-Unif and $\sigma=0.01$ for InfMax-Norm) outperform all baselines in more cases. It is also worth noting that, as suggested by Eq.~(\ref{eq:theta}), the distribution of $\theta$ is dependent on the perturbation $\epsilon$ and hence $\lambda$. In the approximated uniform and normal distributions for InfMax-Unif and InfMax-Norm respectively, the optimal choice of $a$ and $\sigma$ should be dependent on $\lambda$. Intuitively, smaller $\lambda$ makes the $\theta$ have larger variance, so the choice of $a$ and $\sigma$ should also be larger. This is indeed suggested by the results in Figure~\ref{fig:lambda}. Recall that, in Section~\ref{sec:baseline}, we discussed that RWCS can be viewed as a special case of InfMax-Unif with $a=\infty$. And in Figure~\ref{fig:lambda}, we observe that RWCS (equivalent to InfMax-Unif with $a=\infty$) sometimes (e.g., for GCN) outperforms InfMax-Unif (with $a=0.01$) when $\lambda$ is very small. However, we leave further optimization of the hyper-parameters of the proposed strategies to future work. 

\vpara{Sensitivity analysis of $a$ for InfMax-Unif and $\sigma$ for InfMax-Norm.}
In Figure~\ref{fig:sensitivity}, we carry out a sensitivity analysis with resepct to $a$ and $\sigma$ for InfMax-Unif and InfMax-Norm respectively. In Section~\ref{sec:attack-exp}, we have fixed $a=0.01$ and $\sigma=0.01$ for all experiment settings. Here we vary them from 0.005 to 0.02 and show that the results of the proposed strategies, especially those of the InfMax-Norm, stay relatively stable with varying choices of the hyper-parameters.

\subsection{Distributions of $\theta$ on More Nodes}
\label{appendix:hist}
\begin{figure*}
    \centering
    \includegraphics[width=\textwidth]{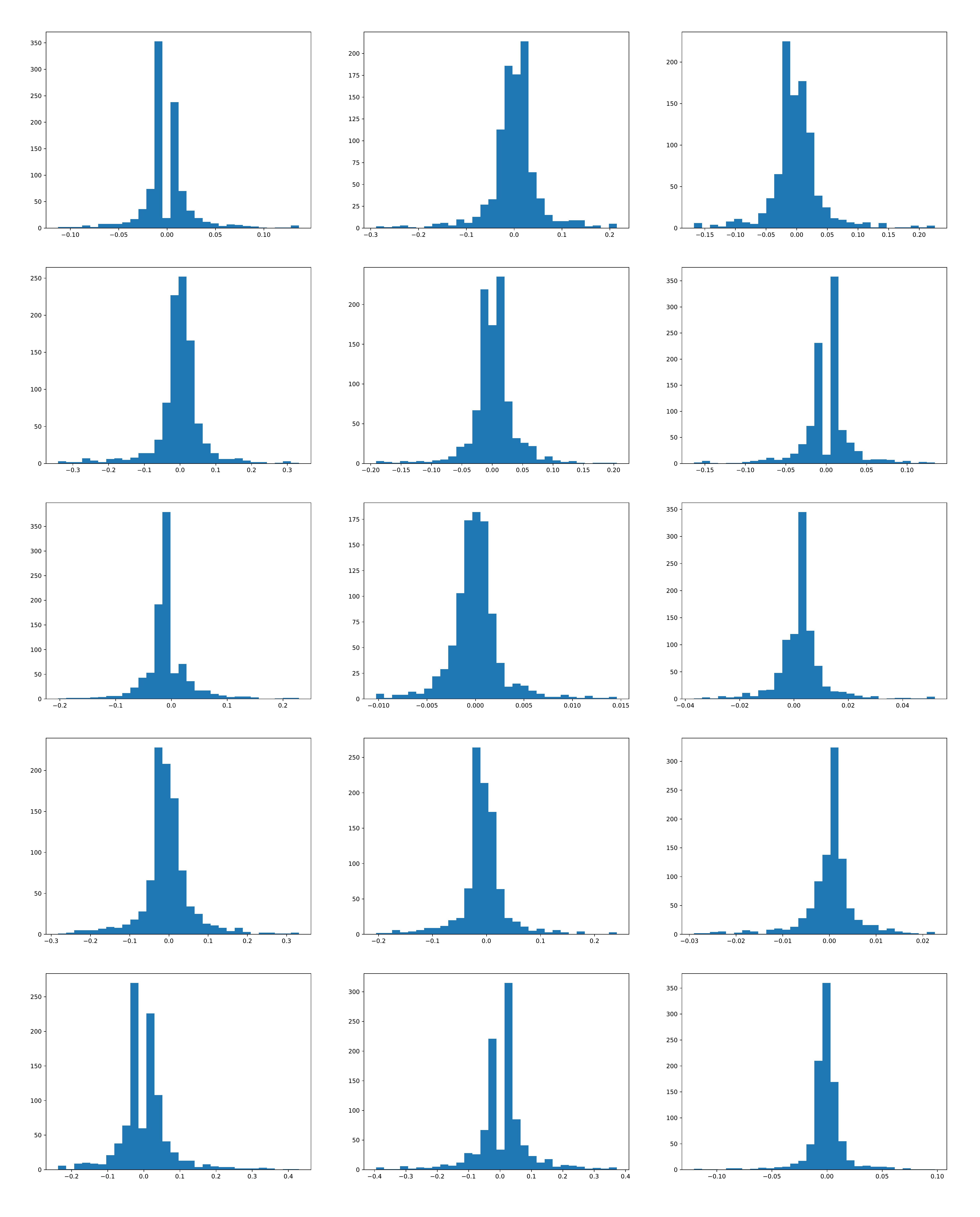}
    \caption{Each figure shows a histogram of $\theta_j$ for a fixed node $j$ over 1000 independent trials of GCN on Cora. The 15 nodes are randomly selected from the union of the validation set and test set.}
    \label{fig:theta15}
\end{figure*}

Distributions of $\theta$ on more randomly selected nodes are provided in Figure~\ref{fig:theta15}. Many examples of the distributions present bell shapes that are close to normal distributions. And it is approximately true that the probability density function is non-increasing at the positive region.

\end{document}